\documentclass{article}
\usepackage{style}
\usepackage[utf8]{inputenc}
\usepackage[style=numeric, backend=biber, maxbibnames=99, minbibnames=1]{biblatex}

\definecolor{gooddarkblue}{rgb}{0,0.1,0.4}
\definecolor{gooddarkred}{rgb}{0.6,0,0.1}
\definecolor{myred}{RGB}{200,0,0}
\definecolor{mygreen}{RGB}{0,130,0}
\hypersetup{        % for better colors
    colorlinks = true,
    citecolor = gooddarkblue,
    linkcolor = gooddarkred,
    filecolor = gooddarkblue,  
    urlcolor = gooddarkblue,  
}

\newcommand{\ouralg}{\textsc{xGAP-Debias}\xspace}
\newcommand{\greenincrease}[1]{\textcolor{mygreen}{#1}}
\newcommand{\reddecrease}[1]{\textcolor{myred}{#1}}

\title{Do Not Harm Protected Groups in Debiasing Language Representation Models}

\author{Chloe Qinyu Zhu \quad\quad Rickard Stureborg \quad\quad Brandon Fain \\
Duke University \\
\texttt{\{}\texttt{qinyu.zhu}, \texttt{rs541}, \texttt{btfain}\texttt{\}}\texttt{@duke.edu}\\
}
\DeclareUnicodeCharacter{FFFD}{\textasciiacute}

\begin{document}

\date{}
\maketitle

\begin{abstract}
  \noindent Language Representation Models (LRMs) trained with real-world data may capture and exacerbate undesired bias and cause unfair treatment of people in various demographic groups. Several techniques have been investigated for applying interventions to LRMs to remove bias in benchmark evaluations on, for example, word embeddings. However, the negative side effects of debiasing interventions are usually not revealed in the downstream tasks. We propose \ouralg, a set of evaluations on assessing the fairness of debiasing. In this work, We examine four debiasing techniques on a real-world text classification task and show that reducing biasing is at the cost of degrading performance for all demographic groups, including those the debiasing techniques aim to protect. We advocate that a debiasing technique should have good downstream performance with the constraint of ensuring no harm to the protected group.
  % \footnote{Code and demo notebooks are available at \url{www.github.com/anonymous/repo}}.
\end{abstract}

\section{Introduction}
Suppose a hiring hospital wants to offer targeted advertisements for an open surgeon position. The employer from the hospital mines the user's bio on social media to predict whether an individual is a surgeon to determine whether to offer the relevant advertisement. To make the prediction, they use a pre-trained Language Representation Model to encode the text and then fine-tune a classification model on top of the representation. The employer decides to use a debiasing technique on the mined data to give equal opportunity to people with different attributes. However, the employer observes that female surgeons receive the advertisement at much lower rates than male surgeons. Even worse, the fraction of female surgeons seeing the advertisement went down after the debiasing intervention. In this work, we show that such scenarios are highly plausible with existing debiasing techniques.

Undesired bias or social stereotypes have been found in natural language representations \cite{bolukbasi2016man}, and systematic ways of debiasing have been widely discussed \cite{gonen-goldberg-2019-lipstick, bender2021dangers, may-etal-2019-measuring, caliskan2017semantics}. Recent works focus on developing techniques to detect, evaluate and mitigate bias in LRMs and reduce harm to marginalized individuals and groups \cite{zmigrod-etal-2019-counterfactual, hardt2016equality,ravfogel-etal-2020-null, may-etal-2019-measuring, caliskan2017semantics, nangia-etal-2020-crows, li-etal-2020-unqovering}. Some of those works measure bias comprising metrics \cite{hardt2016equality, czarnowska-etal-2021-quantifying} and datasets \cite{de2019bias, chalkidis-etal-2022-fairlex} to investigate biases within a specific natural language processing (NLP) task, such as text classification \cite{zhao-etal-2020-gender} or language generation \cite{sheng-etal-2019-woman}. Other works design debiasing techniques for the specific applications in patient notes \cite{minot2022interpretable}, clinical record de-identification \cite{xiao2023name}, or dissecting ML-guided health decisions \cite{obermeyer2019dissecting}.

Due to the variants in datasets and the application area, it is hard to evaluate the downstream performance of the debiasing techniques. Previous work has raised this concern \cite{prost2019debiasing} by utilizing Equality of Opportunity and evaluating the downstream model performance with debiasing across all groups in the dataset. We expand this consideration to examine how debiasing affects group-wise performance and the model performance with other well-known debiasing techniques. 

In this work, we study the effectiveness of the language debiasing technique on the task where the protected attributes are given in the dataset (\Cref{study population}). We propose \ouralg, a framework with a combination of criteria for characterizing fairness in multiple senses: a group-wise utility or performance measure (\textbf{\textsc{x}}) and the corresponding difference of performance \textbf{\textsc{x}} between groups (known as \textbf{GAP}) of model performance on that evaluation metrics between protected attributes. We evaluate four widely-used debiasing techniques on a challenging language model multiclass classification task, where the input is embedded from brief natural language bios, and the target of the classification is the profession. We find that debiasing techniques are either ineffective in reducing the GAP, or are effective at the cost of reducing the model performance on protected attributes, including the group for which debiasing was intended to improve outcomes. In a context where the protected group subject prefers higher model performance, such an intervention achieves `fairness' only through harm.

\section{\ouralg}
\label{satisfaction}
There are many diverse downstream applications for natural language classifiers, and as such, limit to any specific metric will likely have limited utility in some cases. Our \ouralg leaves the flexibility to use any desired evaluation metric to match the use case at hand. The \textsc{x} can, therefore, be any performance evaluation metric used at the class level in the downstream task. 

\textbf{\ouralg Fairness Definition.} We argue that a fair debiasing technique should guarantee that after debiasing:
\begin{enumerate}
  \item Metrics (\textsc{x}) of the protected group\footnote{In this study, we define the `protected group' as the demographic group with a lower performance before applying debiasing techniques.} with respect to the protected attributes should be no worse than before. This can be thought of as a ``do no harm'' criterion.
  \item The \textsc{GAP} of the metrics between protected attributes should decrease substantially. This can be thought of as an ``improvement in equality criterion.''
  \end{enumerate}

  If a debiasing intervention satisfies these two criteria, we consider this \textbf{base satisfaction}. For a multi-class classification problem, we can further break down this criteria satisfaction at the level of individual predictive classes for each demographic group.

  We say an intervention satisfies \textbf{advanced satisfaction} if, in addition to base satisfaction, it does not result in a reduction of performance (measured by \textsc{x}) for the non-protected group(s).

\section{Experiment}
\textbf{Data.}
 In our study, we use \textit{Bias in Bios}\cite{de2019bias}, which contains short online biographies (bios) written in English. Each bio is associated with one of 28 professions and one of two gender identities, where we consider the gender identity as `protected attributes'\footnote{We acknowledge that gender is more complex and non-binary. Following the data collection process in \textit{Bias in Bios}, we used a binary division of gender while investigating the bias in the pre-trained language models and consider the gender group as the only group in all the experiments in this paper.}. We want to predict a profession given the tokenized English bio. Details of the study population can be found in \Cref{study population}. 

\textbf{Problem Statement.}
We want to evaluate the overall and group-wise prediction performance before and after applying each debiasing technique. 

In both the overall and the group-wise evaluation, we consider the True Positive Rate (TPR) of the classification, broken down by gender, as the relevant utility measure \textsc{x} and calculate the difference of TPR between groups. Denote the set of all profession $\mathcal{P}$ for each tokenized bio in the dataset $\mathcal{D}$. For the protected attribute $z$ in all attributes $\mathcal{Z}$, in a given profession $p$, we have binary gender attribute male ($z$) and female ($z'$). The \textsc{GAP} of TPR can be denoted as $\textsc{GAP}^{\text{TPR}}_{z, p}$ with a specific gender $g$ and profession prediction $\hat{p}$.
For a given profession $p$:
\begin{align}
    \textsc{GAP}^{\text{TPR}}_{z, p} = \text{TPR}_{z, p} - \text{TPR}_{z', p}, \quad
    \text{where } \text{TPR}_{z, p} = \mathbb{P}[\hat{P} = p | Z = z, P = p]
\end{align}
where $\hat{P}$, $Z$, and $P$ are predictions for the profession, gender, and ground-truth profession, respectively.

We use the \textsc{GAP} to measure the difference in model performance for the selected evaluation metric between the protected attributes, which quantifies the disparity in the model's classification task performance prediction across different prediction classes and protected attributes.

In the overall performance evaluation, due to the data variance, we involve the idea of \textsc{GAP} Root Mean Square (\textsc{GAP}$^{RMS}$) with TPR to combat the possible impact of data imbalance \cite{ravfogel-etal-2020-null}. We denote the \textsc{GAP}$^{RMS}$ in this experiment as\footnote{For simplicity, in the rest of the paper, \textsc{GAP} refers to TPR \textsc{GAP} and \textsc{GAP}$^{RMS}$ refers to the TPR \textsc{GAP} with Root Mean Square.}:
\begin{align}
    \textsc{GAP}^{\text{RMS}}_{z} = \sqrt{\frac{1}{| \mathcal{P} |}\sum_{\text{p} \in \mathcal{P}}\left(\textsc{GAP}^{\text{TPR}}_{z, \text{p}}\right)^2}
\end{align}
\textsc{GAP}$^{RMS}$ helps to evaluate both the model predictions and the variance with respect to different profession groups. Different from comparing the averaged group-wise TPR overall profession groups, \textsc{GAP}$^{RMS}$ will not be influenced by the imbalanced disparities in TPR in one attribute in one direction. 

We use TPR and Accuracy in conjunction with the evaluation. Accuracy measures the proportion of correctly classified cases overall with the system, regardless of the specific classes the predicted label belongs to. Under extreme data imbalance cases, accuracy might not be an ideal pick of metrics for performance evaluation. Different from the group-wise population, the overall population of the protected attributes is close to each other and accuracy thus turns out to be a valuable metric \textsc{x} for the overall performance evaluation. \Cref{appendixC} elaborates more on evaluation metrics and how they contribute to this experiment.

\textbf{Experimental Setup.} 
In this study, we evaluate the following four debiasing methods (\Cref{appendixA}): 

\begin{minipage}[t]{0.44\linewidth}
        \begin{itemize}
        % \centering
        \small
            \item Equality of Opportunity (EO) \cite{hardt2016equality}
            \item Decoupled Classifiers(Decoupled) \cite{ustun2019fairness}
        \end{itemize}
    \end{minipage}%
\begin{minipage}[t]{0.56\linewidth}
        \begin{itemize}
        \small
            \item Counterfactual Data Augmentation (CDA) \cite{zmigrod-etal-2019-counterfactual}
            \item Iterative Nullspace Projection (INLP) \cite{ravfogel-etal-2020-null}
        \end{itemize}
    \end{minipage}

We use the Logistic Regression classifier with multi-group classification for prediction\footnote{For INLP, we tokenize the English bio with BERT \cite{devlin-etal-2019-bert}}. To validate our results through statistical significance testing, we use the same set of hyperparameters on the classifier and repeat the experiment five times for each debiasing techniques. We report the mean performance across these runs and run a two-sample t-test to investigate if the difference between means is statistically significant. 

\begin{figure}[ht]
  \centering
  \subfloat[Weighted performance regarding group populations]{
    \begin{subfigure}{0.24\linewidth}
        \includegraphics[width=\linewidth]{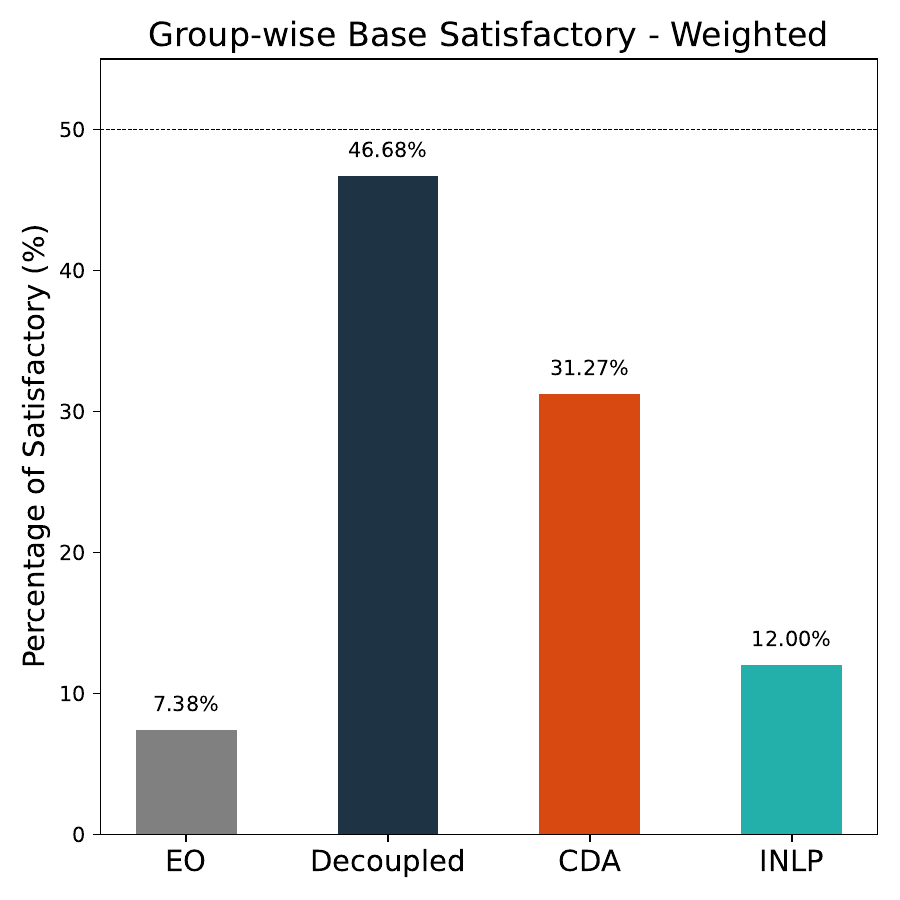}
    \end{subfigure}%
    % \hspace{0.25cm}
    \begin{subfigure}{0.24\linewidth}
        \includegraphics[width=\linewidth]{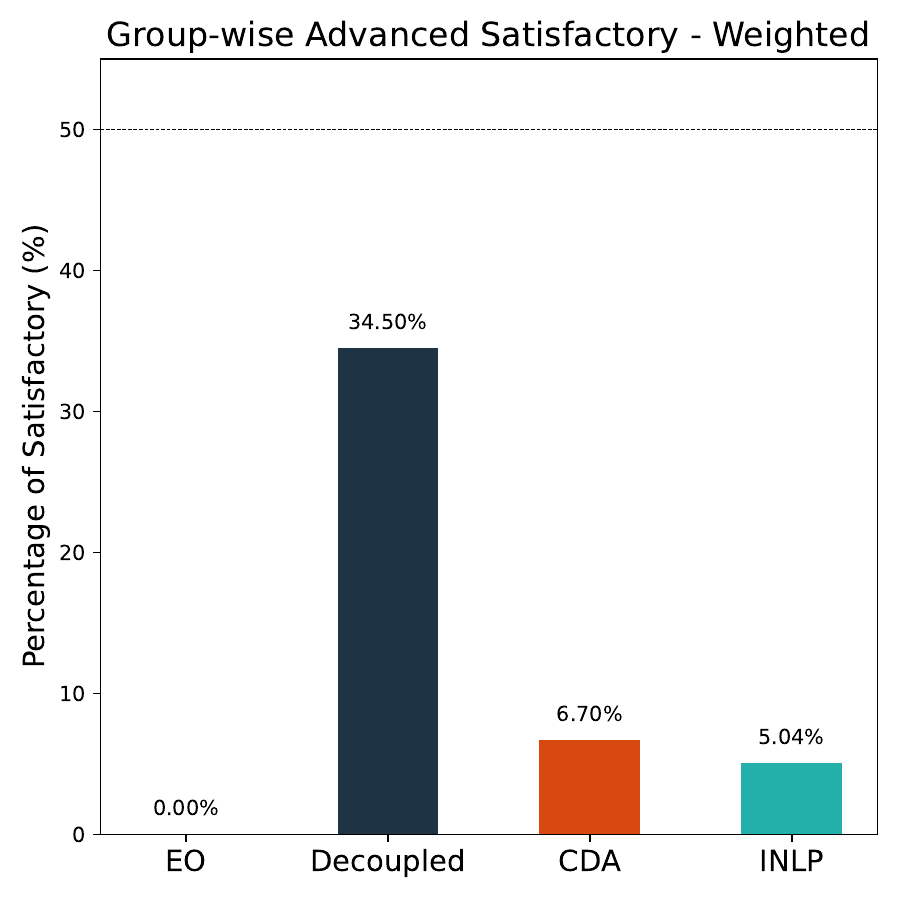}
    \end{subfigure}
    }
   \subfloat[Unweighted performance]{
    % \caption{hi}
    % \captionof{subfigure}{Description for the first pair of figures.}
    \begin{subfigure}{0.24\linewidth}
        \includegraphics[width=\linewidth]{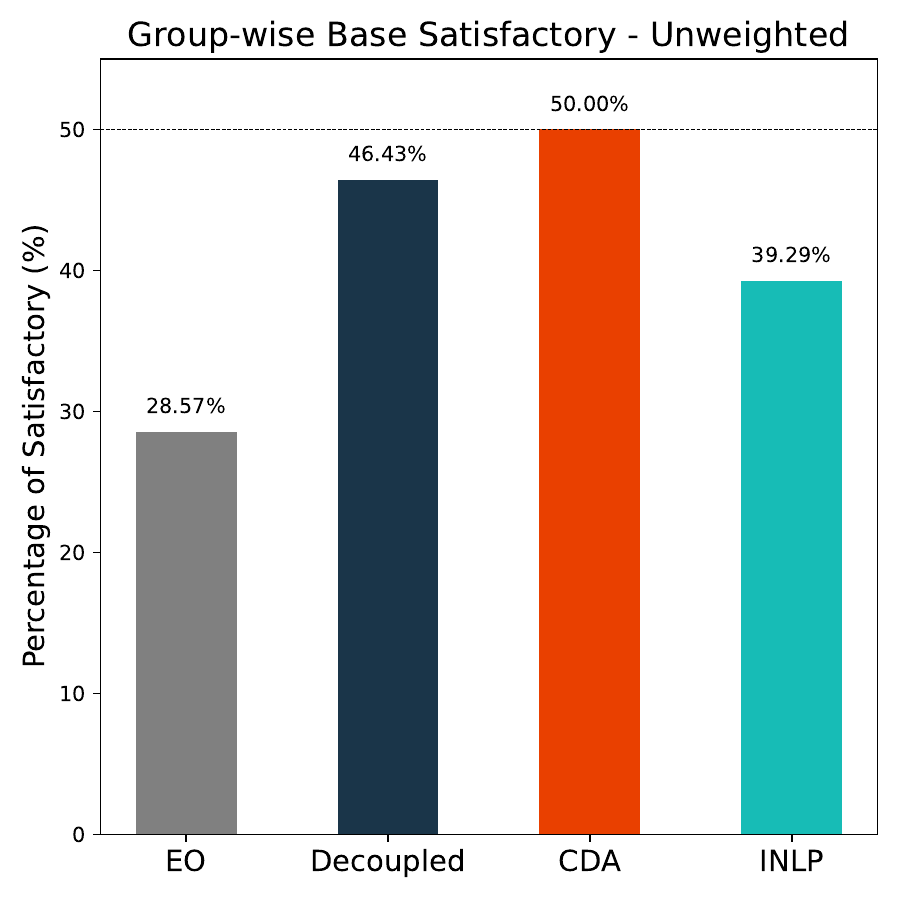}
    \end{subfigure}
    \begin{subfigure}{0.24\linewidth}
        \includegraphics[width=\linewidth]{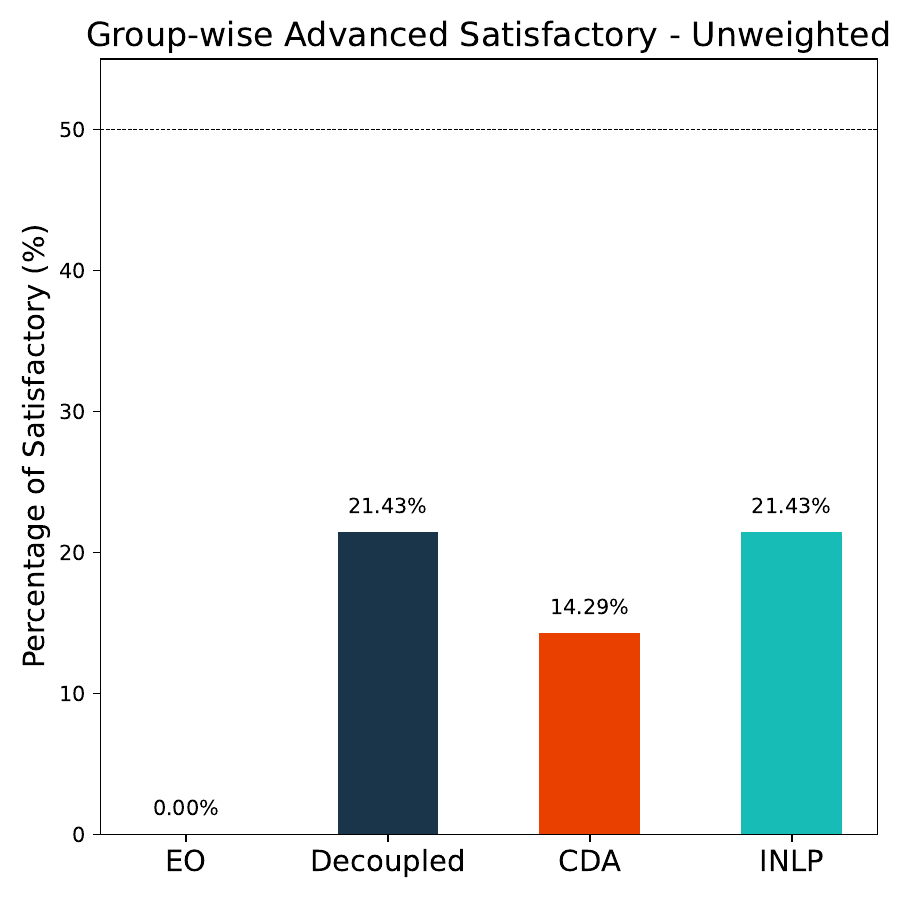}
    \end{subfigure}
  }
  \label{baseadv:a}
  \caption{\textbf{Satisfaction evaluation.} The pair bar at the left-hand side presents the weighted satisfaction rate, while the pair bar at the right-hand side presents the unweighted. After weighting, we observe that INLP has a huge decrease in the total data points that benefited from the debiasing. CDA, which has a higher unweighted satisfaction rate, has impacted only more than half of the Decoupled in the weighted case. However, none of the models under any evaluation criteria has more than 50\% satisfaction.} 
\label{fig_baseadv}
\end{figure}

\textbf{Results.}
% \label{prelim}
\Cref{tab:average_all_model_performance} presents the overall model performance before and after applying the four debiasing techniques. We observe that after debiasing, across all groups, the \textsc{GAP}$^{RMS}$ between the protected attributes decreases substantially. However, this comes at the expense of worsening the model prediction for both groups, including the protected group with worse-off performance before debiasing. The overall performance seems to satisfy the \textsc{GAP} criterion in \ouralg while failing to achieve the first `do no harm' criterion. 

\begin{table*}[ht]
    \caption{\textbf{Overall prediction performance after debiasing.} We evaluate the change of performance for each metric after debiasing with the original performance. We repeated the experiment with five independent model initiations and reported the average means. We then performed a two-sample t-test and underlined any statistically significant values (p-value below 0.05). Note that for TPR and Accuracy, it is considered improved if the prediction score increased after debiasing, whereas a decrease in \textsc{GAP} indicates an improvement. \reddecrease{\textbf{Red}} indicates that a metric has worsened, whereas \greenincrease{\textbf{green}} indicates improved.}
    \label{tab:average_all_model_performance}
    \centering
    % \large
    \begin{tabular}{@{}lcccccc@{}}
        \toprule
        & \multicolumn{1}{c}{Original} & \multicolumn{1}{c}{EO} & \multicolumn{1}{c}{Decoupled} & \multicolumn{1}{c}{CDA} & \multicolumn{1}{c}{INLP} \\
        \midrule
        Accuracy & 79.89 & \underline{72.81} \reddecrease{$\downarrow$7.08} & 79.87 \reddecrease{$\downarrow$0.02} & \underline{79.79} \reddecrease{$\downarrow$0.10} & \underline{78.14} \reddecrease{$\downarrow$1.75} \\
        
        \textsc{GAP}$^{\textit{RMS}}$ & 15.65 & \underline{11.98} \greenincrease{$\downarrow$3.67} & 15.61 \greenincrease{$\downarrow$0.04} & \underline{14.15} \greenincrease{$\downarrow$1.50} & \underline{12.11} \greenincrease{$\downarrow$3.54} \\
        
        TPR$_{\textit{female}}$ & 80.58 & \underline{73.35} \reddecrease{$\downarrow$7.23} & 80.57 \reddecrease{$\downarrow$0.01} & 80.56 \reddecrease{$\downarrow$0.02} & \underline{78.90} \reddecrease{$\downarrow$1.68} \\
        
        TPR$_{\textit{male}}$ & 78.70 & \underline{71.41} \reddecrease{$\downarrow$7.29} & 78.67 \reddecrease{$\downarrow$0.03} & \underline{78.51} \reddecrease{$\downarrow$0.19} & \underline{76.77} \reddecrease{$\downarrow$1.93} \\
        \bottomrule
    \end{tabular}
\end{table*}

\Cref{appendixA} details the performance after debiasing on each of the 28 professions for all four techniques. Among all the professions, INLP has more than 85\% groups where the \textsc{GAP} decreases after debiasing (\Cref{inlp-evaluation}). Meanwhile, the change of \textsc{GAP} with INLP is also the best among all the debiasing techniques (\Cref{gap change}). We detail the change of TPR regarding the protected attributes in \Cref{appendixC}. To address the impact of data imbalance, we consider both unweighted and weighted calculations with respect to the group population in evaluating the group-wise satisfaction rate among different professions (the prediction groups). \Cref{fig_baseadv} shows base and advanced satisfaction with weighted and unweighted calculations. The weighted performance of the debiasing techniques is more distinctive than the unweighted. While the total number of satisfaction is similar, one technique might have the satisfaction on the profession with a large population. Decoupled Classifiers constantly outperform the other debiasing techniques, regardless of whether take total affected data points into consideration. Unfortunately, for all four debiasing techniques, none of them exceed 50\% satisfaction rate under both base and advanced satisfaction criteria.
\begin{wraptable}{r}{0.5\linewidth}
    \centering
    \vspace{+5pt}
    \caption{\textbf{Percentage of professions with worsened \textsc{GAP} metrics after debiasing}.}
    \vspace{-5pt}
    \label{tab:percentage_worsened_professions}
    \begin{tabular}{@{}p{0.45\linewidth}|p{0.35\linewidth}}
        \toprule
        Method & Worsened \textsc{GAP} \\
        \midrule
        EO & 39\% \\
        Decoupled & 39\% \\
        CDA & 39\% \\
        INLP & 14\% \\
        \bottomrule
    \end{tabular}
    \vspace{-15pt}
\end{wraptable}

\Cref{gap change} shows the changes in \textsc{GAP} broken down by profession. Note that while EO had the greatest improvement in \textsc{GAP} (\Cref{tab:average_all_model_performance}), that performance increase is not equally spread across all professions.
Likewise, \Cref{tab:percentage_worsened_professions} shows the percentage of professions that worsened in the \textsc{GAP} metric for each debiasing method.
We observe that no method is able to achieve an overall reduction in \textsc{GAP} without increasing the \textsc{GAP} in a sizeable portion of the professions. In particular, EO, Decoupled, and CDA increase the \textsc{GAP} in more than a third of professions.
This adds a new dimension of complexity to the analysis of reducing harm, which has not yet been explored.

\section{Discussion}
Our work introduces \ouralg, a framework to evaluate the effectiveness and fairness of language debiasing models on specific downstream tasks. From the multiclass classification task, we investigate the fairness of debiasing models in different perspectives: we introduce base and advance \ouralg rule of satisfaction, evaluate both overall and group-wise performance (\Cref{tab:average_all_model_performance}, \Cref{appendixA}), compare the weighted and unweighted satisfactions (\Cref{fig_baseadv}), and measured the proportion of groups that are harmed by increasing the overall performance (\Cref{tab:percentage_worsened_professions}). From the result, we found that none of the debiasing models achieves over 50\% \ouralg satisfaction.

\begin{figure}[ht]
    \centering    
    \includegraphics[width=0.8\linewidth]{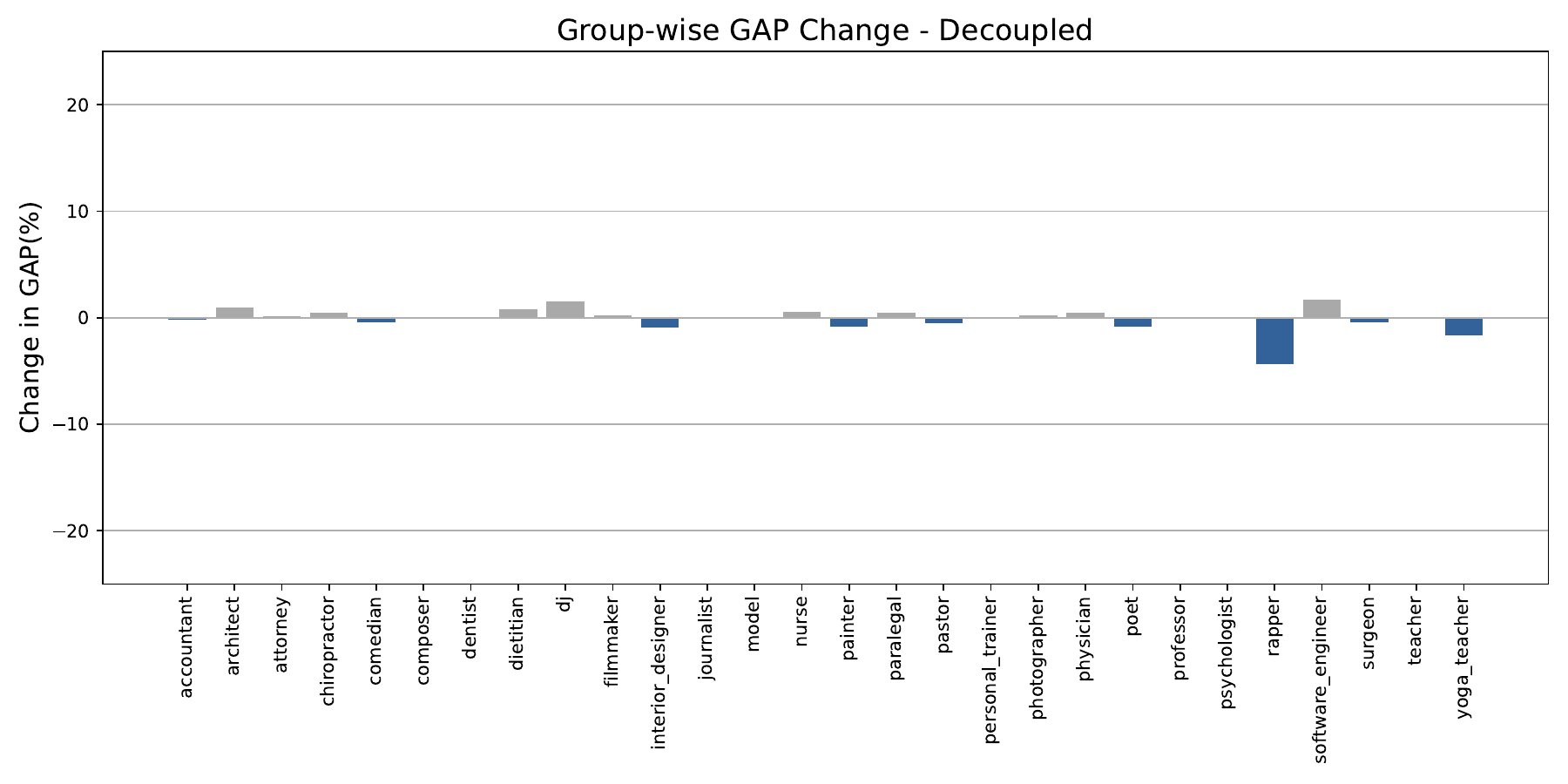}
    \caption{\textbf{Scale of changes in GAP across professions - Decoupled.} We compare the TPR GAP before and after applying debiasing techniques. If the GAP shrinks between demographic groups for the profession, the corresponding bar would be highlighted in \textcolor{gooddarkblue}{\textbf{blue}}.}
    \label{gap_change_decoupled}
\end{figure}

\begin{figure}[ht]
    \centering    
    \includegraphics[width=0.8\linewidth]{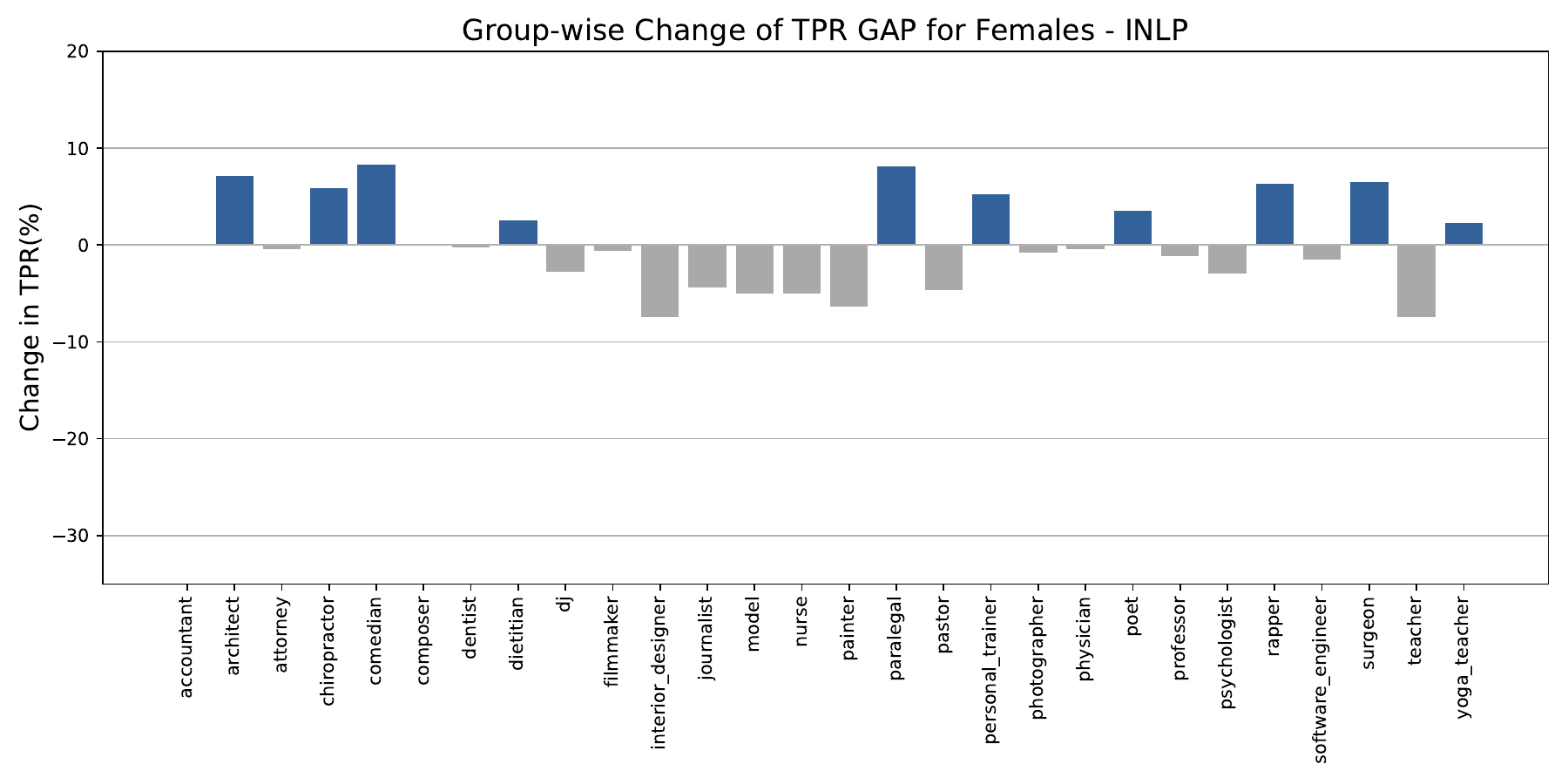}
    \caption{\textbf{Scale of changes in true positive rate across professions - INLP.} We evaluate the change of TPR before and after applying INLP debiasing technique for female in each of the professions. We highlight the improved profession in \textcolor{gooddarkblue}{\textbf{blue}}.}
    \label{tpr_female_change_inlp}
\end{figure}

Among the four models, the Decoupled seems to have the best performance in satisfaction results. However, as we delve deep into the results of \ouralg metrics, we found that in group-wise evaluation, the Decoupled does not have much improvements in reducing the \textsc{GAP} or increasing the TPR for the protected group (\Cref{gap_change_decoupled}, \Cref{female change}). And INLP worsen about half of the professions in the TPR performance for the protected group (\Cref{tpr_female_change_inlp}). When examining the effects of these debiasing techniques, we see a further complexity not yet addressed in the literature. While all techniques cause a decrease in \textsc{GAP} metrics, this effect is not uniform across professions. As stated in \cite{ravfogel-etal-2020-null}, INLP is known for reducing \textsc{GAP} after debiasing. When we move one step deeper, from the overall performance to the group-wise performance, INLP hurts more than one-third of the groups to achieve a reduction of \textsc{GAP} in the report of overall performance (\Cref{tab:percentage_worsened_professions}, \Cref{gap change}). 

In the previous discussion, one might found that the model with high \ouralg satisfaction is actually not doing the debiasing job well. One might question how \ouralg assists in evaluating fair and effective language debiasing techniques. We advocate that \ouralg should be considered as the first constraint in the debiasing evaluation. It means a debiasing model should guarantee to maintain a high satisfaction rate with \ouralg before reaching the goal of improving the prediction performance. Therefore, a good debiasing technique \textit{should} consider \ouralg as to pass an `entry test' to demonstrate the model robustness in fairness without harm to the protected group(s).

\section{Conclusion and Future Work}
In our study, we have highlighted difficulties with existing debiasing techniques when used as part of an intervention on a language classification task. Through \ouralg, we highlight practical challenges that existing debiasing techniques face to remain `fair' after debiasing in the downstream applications, such as a tradeoff in performance or imbalanced changes in performance across the target classes.
There is a gap in the current state of the art for a principled debiasing technique that can guarantee higher satisfaction rates of both our `do no harm' and `improvement of equality' criteria.
Our evaluation motivates the need for multiple assessments of fairness to ensure bias reduction without harm.

%We noted that in the evaluation of scale changes, Decoupled Classifiers, which has the best satisfaction rate, does not outperform the rest of the techniques. On the other hand, Decoupled Classifiers mainly hold the model performance before debiasing with slight improvement. In our \ouralg, we did not have a strict cut-off line for the scale of changes. Due to the variety of downstream applications and the limit of publically available large multiclassification datasets, we cannot access the dataset in a specified field such as clinical treatment or financial loan. Besides raising social awareness on the fairness of debiasing evaluation, we does not propose a strong preliminary result that can be used as a baseline as a further step of development. Future work that applies our \ouralg under a specific downstream task might consider quantifying the scale of changes in the fairness of evaluation with the area of knowledge, evaluating \ouralg on the domain-specific datasets, and improving current debiasing techniques.

\section{Social Impact Statement}
% 1> put in the social impact saying that it can benefit the disabled people: https://arxiv.org/abs/2005.00813
% 2> 
The goal of debiasing is to intentionally tip the scale back in favor of those who face discrimination without inadvertently perpetuating harm or injustice. 
As pointed back to the hiring story, one may wish to debias the model toward the direction that the gap between men and women is reduced. However, the inadvertent effect might be that our model becomes an even poorer recruitment tool for finding surgeons, sacrificing the model performance for all protected attributes to have a closer prediction performance, even harming the very group we intended to increase equity. With the development of language technology, it is unavoidable to observe bias within the model. We are committed to advancing the research on reducing bias in language models, which requires more robust evaluations and frameworks for fairness.

\newpage
\printbibliography
\appendix

\newpage
\section{Profession-Level Model Performance}
\label{appendixA}
We provide more details about group-wise model performance for each of the debiasing technique with \ouralg. We underline the profession that achieves base satisfaction and the profession underlined with a star($^*$) refers to achieving advanced satisfaction. In the parenthesis, we include the standard deviation from repeating the experiment five times.

\subsection{Equality of Opportunity}
Equality of Opportunity is a measure of debiasing discrimination. If the TPR of two protected attributes is the same, they are equally qualified for a positive output. They should have the same probability of being correctly classified by the language model. By optimizing accuracy, the classifier can meanwhile optimize a form of Equality of Opportunit (EO). Then we measure the cost of the EO, ensuring equality of opportunity with respect to the accuracy \cite{hardt2016equality}.

\begin{table*}[ht]
    \caption{\textbf{Group-wise Equality of Opportunity model performance}}
    \label{eqodds-evaluation}
    \centering
    % \small
    \captionsetup{font=footnotesize}
    \begin{adjustbox}{width=\textwidth}
    \begin{tabular}{@{}lrrrrrr@{}}
    \toprule
     & \multicolumn{2}{c}{TPR Male} &   \multicolumn{2}{c}{TPR Female} & \multicolumn{2}{c}{GAP}\\
     \cmidrule(lr){2-3} \cmidrule(l){4-5} \cmidrule(l) {6-7}
     {Profession} & \multicolumn{1}{c}{Original} & \multicolumn{1}{c}{Debiased} & \multicolumn{1}{c}{Original} & \multicolumn{1}{c}{Debiased} & \multicolumn{1}{c}{Original} & \multicolumn{1}{c}{Debiased} \\
     \midrule
     Accountant & 64.77 (0.004) &56.85 (0.008) &63.45 (0.004) & 57.58 (0.006) & 1.32 (0.003) & \textbf{0.76} (0.006)\\
     Architect & 55.91 (0.016) & 38.02 (0.013) & 57.79 (0.012) &50.99 (0.003)  & 1.89 (0.009)& 12.97 (0.015) \\
     Attorney & 87.21 (0.001) & 82.58 (0.004) & 84.82 (0.004) & 81.60 (0.009)& 2.39 (0.003) &\textbf{0.99} (0.006) \\
     \underline{Chiropractor} & 28.19 (0.004) & 23.79 (0.003)& 20.00 (0.004) & \textbf{21.23} (0.009) & 8.19 (0.004) & \textbf{2.56} (0.008) \\
     \underline{Comedian} & 78.30 (0.006) &74.13 (0.003) &62.31 (0.003) &\textbf{66.46} (0.014) &15.99 (0.004) &\textbf{7.67} (0.012)\\
     Composer & 89.46 (0.002) & 83.73 (0.016) & 84.31 (0.003) & 77.87 (0.005) & 5.15 (0.001) &5.86 (0.019) \\
     Dentist & 89.91 (0.001) & 87.39 (0.001) & 92.21 (0.001) & 91.36 (0.001) & 2.30 (0.002) & 3.97 (0.002) \\
     \underline{Dietitian} & 56.25 (0.014) & \textbf{64.21} (0.025) & 76.20 (0.006) & 58.15 (0.019) & 19.95 (0.011) & \textbf{6.06} (0.020)\\
     \underline{Dj} & 75.53 (0.011) &67.25 (0.029)  & 46.25 (0.014) &\textbf{57.50} (0.031)  &29.28 (0.012)  & \textbf{9.75} (0.057)\\
     Filmmaker & 80.95 (0.004) & 75.75 (0.008) & 76.30 (0.006) & 73.89 (0.004) & 4.64 (0.004) & \textbf{1.86} (0.012)\\
     \underline{Interior Designer}& 44.23 (0.018) &\textbf{48.31} (0.034) & 68.67 (0.007) & 57.40 (0.012) & 24.44 (0.012) &\textbf{9.09} (0.031) \\
     Journalist & 76.13 (0.003) & 64.28 (0.014) & 78.29 (0.004) & 63.42 (0.009) & 2.16 (0.003)& \textbf{1.70} (0.013)\\
     Model & 42.65 (0.002) & 39.04 (0.007) &81.35 (0.002)  & 65.86 (0.003) & 38.70 (0.002) & \textbf{26.83} (0.009)\\
     Nurse & 70.42 (0.004) & 69.90 (0.003) & 80.54 (0.005) & 69.91 (0.003) & 10.12 (0.006) & \textbf{0.15} (0.001)\\
     Painter & 77.47 (0.006) & 71.44 (0.011) & 79.39 (0.001) & 71.87 (0.023) & 1.92 (0.005) & 2.45 (0.010)\\
     \underline{Paralegal} & 6.45 (0.000) & \textbf{24.19} (0.059) & 35.29 (0.021) & 11.66 (0.003) & 28.84 (0.021) & \textbf{12.53} (0.060)\\
     Pastor & 62.09 (0.004) &32.06 (0.016)  & 55.78 (0.010) &50.34 (0.033) & 6.31 (0.013) & 18.28 (0.039)\\
     Personal Trainer & 72.45 (0.009) &54.80 (0.016)  & 60.80 (0.012) & 60.49 (0.006) & 11.65 (0.011)& \textbf{5.70} (0.021)\\
     Photographer & 87.33 (0.001) & 79.42 (0.006) &84.25 (0.002) & 79.92 (0.008) & 3.08 (0.001) &\textbf{0.77} (0.003) \\
     Physician & 78.45 (0.008) & 70.49 (0.001) & 95.20 (0.001) & 93.87 (0.001) & 16.75 (0.007) & 23.38 (0.001)\\
     Poet & 74.92 (0.005) & 70.83 (0.008) & 76.44 (0.004) & 69.56 (0.017) & 1.52 (0.002) &2.34 (0.010) \\
     Professor & 87.79 (0.003) & 82.96 (0.001) & 87.98 (0.002) & 85.09 (0.003) & 0.20 (0.001) & 2.13 (0.003)\\
     Psychologist & 62.30 (0.008) &\textbf{73.71} (0.002) & 65.32 (0.007) & 48.38 (0.014) & 3.02 (0.002)& 25.33 (0.012) \\
     Rapper & 84.07 (0.011) &80.49 (0.014)  & 68.48 (0.031) & 58.26 (0.024) & 15.59 (0.023) & 22.23 (0.025)\\
     Software Engineer & 72.19 (0.014) & 51.84 (0.027) & 68.02 (0.022) &\textbf{72.89} (0.010)&4.17 (0.021) & 21.06 (0.031) \\
     \underline{Surgeon} & 62.36 (0.007) & 40.18 (0.008) & 36.12 (0.004) &\textbf{36.93} (0.006) & 26.24 (0.004) & \textbf{3.25} (0.011)\\
     Teacher & 45.39 (0.010) & 34.53 (0.008) & 58.98 (0.008) & 36.65 (0.009)& 13.59 (0.003)&\textbf{2.12} (0.009) \\
     \underline{Yoga Teacher} & 43.36 (0.020) & \textbf{55.63} (0.009) & 67.59 (0.010) & 54.59 (0.014) & 24.23 (0.022) & \textbf{1.96} (0.011) \\
    \bottomrule
    \end{tabular}
    \end{adjustbox}

\end{table*}

\subsection{Decoupled Classifiers} 
Decoupled Classifiers involve training multiple classifiers independently for different protected attributes \cite{ustun2019fairness}. By training separately, each classifier can concentrate on learning the pattern specific to the group, which can improve the performance and accuracy of each individual classifier. Also, since each classifier is independent, the model has the flexibility to add or modify specific groups without affecting the rest, which is adaptive for applications with dynamic datasets. 

\begin{table*}[ht]
    \caption{\textbf{Group-wise Decoupled Classifiers model performance} }
    \label{decoupled-evaluation}
    \centering
    \small
    \captionsetup{font=footnotesize}
    \begin{adjustbox}{width=\textwidth}
    \begin{tabular}{@{}lrrrrrr@{}}
    \toprule
     & \multicolumn{2}{c}{TPR Male} &   \multicolumn{2}{c}{TPR Female} & \multicolumn{2}{c}{GAP}\\
     \cmidrule(lr){2-3} \cmidrule(l){4-5} \cmidrule(l) {6-7}
     {Profession} & \multicolumn{1}{c}{Original} & \multicolumn{1}{c}{Debiased} & \multicolumn{1}{c}{Original} & \multicolumn{1}{c}{Debiased} & \multicolumn{1}{c}{Original} & \multicolumn{1}{c}{Debiased} \\
     \midrule
     \underline{Accountant} & 64.77 (0.004) & 64.74 (0.003)&63.45 (0.004) &  \textbf{63.64} (0.001)& 1.32 (0.003) & \textbf{1.10} (0.003)\\
     Architect & 55.91 (0.016) & 54.28 (0.019) & 57.79 (0.012) & 57.14 (0.016) & 1.89 (0.009)&2.85 (0.004)  \\
     Attorney & 87.21 (0.001) & 87.15 (0.002) & 84.82 (0.004) & 84.61 (0.001)& 2.39 (0.003) & 2.54 (0.001)\\
     Chiropractor & 28.19 (0.004) & \textbf{28.33} (0.003)& 20.00 (0.004) & 19.69 (0.003) & 8.19 (0.004) & 8.63 (0.003) \\
     \underline{Comedian} & 78.30 (0.006) &78.00 (0.005) &62.31 (0.003) & \textbf{62.46} (0.003)&15.99 (0.004) &\textbf{15.53} (0.007)\\
     \underline{Composer}* & 89.46 (0.002) & \textbf{89.49} (0.002) & 84.31 (0.003) & \textbf{84.47} (0.002) & 5.15 (0.001) & \textbf{5.02} (0.003)\\
     \underline{Dentist} & 89.91 (0.001) & \textbf{89.99} (0.002) & 92.21 (0.001) & 92.19 (0.001) & 2.30 (0.002) & \textbf{2.21} (0.003) \\
     Dietitian & 56.25 (0.014) & 54.47 (0.055) & 76.20 (0.006) & 75.23 (0.028) & 19.95 (0.011) & 20.76 (0.028)\\
     Dj & 75.53 (0.011) & 74.80 (0.005) & 46.25 (0.014) & 44.00 (0.034) &29.28 (0.012)  & 30.80 (0.031)\\
     Filmmaker & 80.95 (0.004) & \textbf{81.09} (0.004) & 76.30 (0.006) &76.24 (0.009)  & 4.64 (0.004) &4.85 (0.006) \\
     \underline{Interior Designer}& 44.23 (0.018) & \textbf{44.62} (0.019) & 68.67 (0.007) & 68.13 (0.007) & 24.44 (0.012) & \textbf{23.52} (0.013)\\
     Journalist & 76.13 (0.003) & 75.55 (0.006) & 78.29 (0.004) & 77.63 (0.007) & 2.16 (0.003)& \textbf{2.08} (0.002)\\
     \underline{Model}* & 42.65 (0.002) & \textbf{42.75} (0.004) &81.35 (0.002)  & \textbf{81.42} (0.006) & 38.70 (0.002) & \textbf{38.68} (0.006)\\
     Nurse & 70.42 (0.004) & 70.40 (0.005) & 80.54 (0.005) & \textbf{81.10} (0.007) & 10.12 (0.006) & 10.70 (0.004)\\
     \underline{Painter} & 77.47 (0.006) & \textbf{77.63} (0.007) & 79.39 (0.001) & 78.74 (0.007) & 1.92 (0.005) & \textbf{1.11} (0.004)\\
     Paralegal & 6.45 (0.000) & \textbf{6.45} (0.000) & 35.29 (0.021) & \textbf{35.73} (0.011) & 28.84 (0.021) & 29.27 (0.011)\\
     \underline{Pastor} & 62.09 (0.004) & 61.56 (0.008) & 55.78 (0.010) & \textbf{55.78} (0.011) & 6.31 (0.013) & \textbf{5.78} (0.005)\\
     \underline{Personal Trainer}* & 72.45 (0.009) & \textbf{72.45} (0.004) & 60.80 (0.012) & \textbf{60.86} (0.009) & 11.65 (0.011)&\textbf{11.58} (0.010) \\
     Photographer & 87.33 (0.001) & \textbf{87.34} (0.002) &84.25 (0.002) & 84.05 (0.004) & 3.08 (0.001) & 3.29 (0.002)\\
     Physician & 78.45 (0.008) & 77.86 (0.007) & 95.20 (0.001) & 95.04 (0.001) & 16.75 (0.007) & 17.18 (0.006)\\
     Poet & 74.92 (0.005) & 74.37 (0.009) & 76.44 (0.004) & 75.04 (0.010) & 1.52 (0.002) & \textbf{0.67} (0.003)\\
     \underline{Professor}* & 87.79 (0.003) & \textbf{88.29} (0.001) & 87.98 (0.002) & \textbf{88.49} (0.002) & 0.20 (0.001) & \textbf{0.19} (0.001)\\
     Psychologist & 62.30 (0.008) &61.89 (0.008) & 65.32 (0.007) & 64.79 (0.008) & 3.02 (0.002)& \textbf{2.90} (0.002) \\
     \underline{Rapper}* & 84.07 (0.011) & \textbf{84.27} (0.005) & 68.48 (0.031) & \textbf{73.04} (0.036) & 15.59 (0.023) & \textbf{11.22} (0.037) \\
     Software Engineer & 72.19 (0.014) & \textbf{73.05} (0.008) & 68.02 (0.022) & 67.21 (0.009) &4.17 (0.021) & 5.85 (0.006) \\
     \underline{Surgeon} & 62.36 (0.007) & 61.98 (0.010) & 36.12 (0.004) & \textbf{36.13} (0.009) & 26.24 (0.004) & \textbf{25.84} (0.004)\\
     Teacher & 45.39 (0.010) & 44.79 (0.007) & 58.98 (0.008) & 58.27 (0.009) & 13.59 (0.003)& \textbf{13.47} (0.003)\\
     \underline{Yoga Teacher}* & 43.36 (0.020) & \textbf{45.31} (0.025) & 67.59 (0.010) & \textbf{67.86} (0.008) & 24.23 (0.022) &  \textbf{22.55} (0.021)\\
    \bottomrule
    \end{tabular}
    \end{adjustbox}

\end{table*}

\subsection{Counterfactual Data Augmentation}
Counterfactual Data Augmentation (CDA) is a technique that debiases by adjusting the training dataset \cite{zmigrod-etal-2019-counterfactual}. As counterfactual reasoning, CDA generates counterfactual instances with respect to the protected attributes. For example, if we have binary gender as a protected attribute, by doing data augmentation, the sentence 'she is happy' would be augmented with 'he is happy.' We consulted \cite{zhao-etal-2018-gender} for the full pair of words list used in our experiment.

\begin{table*}[ht]
    \caption{\textbf{Group-wise Counterfactual Data Augmentation model performance }}
    \label{cda-evaluation}
    \centering
    \small
    \captionsetup{font=footnotesize}
    \begin{adjustbox}{width=\textwidth}
    \begin{tabular}{@{}lrrrrrr@{}}
    \toprule
     & \multicolumn{2}{c}{TPR Male} &   \multicolumn{2}{c}{TPR Female} & \multicolumn{2}{c}{GAP}\\
     \cmidrule(lr){2-3} \cmidrule(l){4-5} \cmidrule(l) {6-7}
     Profession & \multicolumn{1}{c}{Original} & \multicolumn{1}{c}{Debiased} & \multicolumn{1}{c}{Original} & \multicolumn{1}{c}{Debiased} & \multicolumn{1}{c}{Original} & \multicolumn{1}{c}{Debiased} \\
     \midrule
     Accountant & 64.77 (0.004) & 64.31 (0.009) &63.45 (0.004) & 62.75 (0.007)  & 1.32 (0.003) & 1.56 (0.003)\\
     Architect & 55.91 (0.016) & 53.79 (0.013)  & 57.79 (0.012) & 57.04 (0.003) & 1.89 (0.009)& 3.26 (0.012) \\
     \underline{Attorney} & 87.21 (0.001) & 86.97 (0.004) & 84.82 (0.004) & \textbf{84.82} (0.007) & 2.39 (0.003) &\textbf{2.15} (0.002) \\
     Chiropractor & 28.19 (0.004) & \textbf{29.65} (0.014)& 20.00 (0.004) & \textbf{21.33} (0.014) & 8.19 (0.004) & 8.31 (0.008) \\
     Comedian & 78.30 (0.006) & 77.96 (0.003)&62.31 (0.003) & 61.08 (0.007)&15.99 (0.004) & 16.88 (0.010)\\
     \underline{Composer} & 89.46 (0.002) & 88.80 (0.003) & 84.31 (0.003) & \textbf{84.79} (0.007) & 5.15 (0.001) & \textbf{4.01} (0.005)\\
     Dentist & 89.91 (0.001) & \textbf{90.03} (0.001) & 92.21 (0.001) & \textbf{92.43} (0.002) & 2.30 (0.002) &  2.40 (0.002)\\
     \underline{Dietitian}* & 56.25 (0.014) & \textbf{62.11} (0.011) & 76.20 (0.006) & \textbf{78.71} (0.005) & 19.95 (0.011) & \textbf{16.61} (0.008)\\
     \underline{Dj} & 75.53 (0.011) & 75.17 (0.004) & 46.25 (0.014) & \textbf{46.50} (0.022) &29.28 (0.012)  & \textbf{28.67} (0.021)\\
     Filmmaker & 80.95 (0.004) & \textbf{81.38} (0.002) & 76.30 (0.006) & \textbf{76.31} (0.008) & 4.64 (0.004) & 5.07 (0.007)\\
     \underline{Interior Designer}& 44.23 (0.018) & \textbf{52.00} (0.007) & 68.67 (0.007) & 67.73 (0.004) & 24.44 (0.012) & \textbf{15.73} (0.006)\\
     Journalist & 76.13 (0.003) & 75.58 (0.007) & 78.29 (0.004) & 77.24 (0.011) & 2.16 (0.003)& \textbf{1.66} (0.006)\\
     \underline{Model} & 42.65 (0.002) & \textbf{44.19} (0.010) &81.35 (0.002)  & 81.04 (0.003) & 38.70 (0.002) & \textbf{36.85} (0.007)\\
     \underline{Nurse}* & 70.42 (0.004) & \textbf{71.48} (0.005) & 80.54 (0.005) & \textbf{80.78} (0.008) & 10.12 (0.006) & \textbf{9.00} (0.003)\\
     Painter & 77.47 (0.006) & 76.71 (0.004) & 79.39 (0.001) & 79.01 (0.004) & 1.92 (0.005) & 2.31 (0.005)\\
     \underline{Paralegal}* & 6.45 (0.000) & \textbf{11.94} (0.009) & 35.29 (0.021) & \textbf{39.16} (0.013) & 28.84 (0.021) & \textbf{27.22} (0.012)\\
     \underline{Pastor} & 62.09 (0.004) & 60.49 (0.008) & 55.78 (0.010) & \textbf{56.73} (0.006)& 6.31 (0.013) & \textbf{3.76} (0.010) \\
     \underline{Personal Trainer} & 72.45 (0.009) & 69.08 (0.005) & 60.80 (0.012) & \textbf{61.11} (0.006) & 11.65 (0.011)& \textbf{7.97} (0.010)\\
     \underline{Photographer} & 87.33 (0.001) & 86.87 (0.002) &84.25 (0.002) & \textbf{84.41} (0.004) & 3.08 (0.001) & \textbf{2.46} (0.002)\\
     Physician & 78.45 (0.008) & 77.19 (0.008) & 95.20 (0.001) & 95.02 (0.002) & 16.75 (0.007) & 17.83 (0.007)\\
     Poet & 74.92 (0.005) & 74.29 (0.006) & 76.44 (0.004) & \textbf{76.51} (0.006) & 1.52 (0.002) & 2.23 (0.002)\\
     Professor & 87.79 (0.003) & \textbf{88.48} (0.007) & 87.98 (0.002) & \textbf{88.36} (0.009) & 0.20 (0.001) & 0.25 (0.001)\\
     \underline{Psychologist} & 62.30 (0.008) & \textbf{62.44} (0.009)& 65.32 (0.007) & 64.29 (0.012) & 3.02 (0.002)& \textbf{1.85} (0.004) \\
     \underline{Rapper} & 84.07 (0.011) & 83.90 (0.004) & 68.48 (0.031) & \textbf{82.61} (0.000) & 15.59 (0.023) & \textbf{1.29} (0.004)\\
     Software Engineer & 72.19 (0.014) & \textbf{72.27} (0.016) & 68.02 (0.022) & 67.51 (0.019)  &4.17 (0.021) & 4.76 (0.006) \\
     Surgeon & 62.36 (0.007) & 60.54 (0.008) & 36.12 (0.004) & 35.37 (0.005)& 26.24 (0.004) & \textbf{25.17} (0.005)\\
     Teacher & 45.39 (0.010) & 44.21 (0.013) & 58.98 (0.008) & 57.44 (0.012)& 13.59 (0.003)& \textbf{13.23} (0.003)\\
     \underline{Yoga Teacher}* & 43.36 (0.020) & \textbf{49.06} (0.009) & 67.59 (0.010) & \textbf{69.06} (0.006) & 24.23 (0.022) & \textbf{20.00} (0.008) \\
    \bottomrule
    \end{tabular}
    \end{adjustbox}

\end{table*}

\subsection{Iterative Nullspace Projection}

Iterative Nullspace Projection (INLP) works by iteratively projecting the feature vectors of a pre-trained language model onto a subspace that is orthogonal to the subspace spanned by the protected attributes, effectively 'nulling out' the protected attributes from the feature representation  \cite{ravfogel-etal-2020-null}. By doing so, the model is forced to focus on other relevant features that are not correlated with the protected attributes. The INLP algorithm iteratively projects the feature vectors onto the orthogonal space of the subspace spanned by the protected attributes until the resulting feature representation is orthogonal to the protected attributes.

\begin{table*}[ht]

    \caption{\textbf{Group-wise INLP model performance}}
    \label{inlp-evaluation}
    \centering
    \small
    \captionsetup{font=footnotesize}
    \begin{adjustbox}{width=\textwidth}
    \begin{tabular}{@{}lrrrrrr@{}}
    \toprule
     & \multicolumn{2}{c}{TPR Male} &   \multicolumn{2}{c}{TPR Female} & \multicolumn{2}{c}{GAP}\\
     \cmidrule(lr){2-3} \cmidrule(l){4-5} \cmidrule(l) {6-7}
     {Profession} & \multicolumn{1}{c}{Original} & \multicolumn{1}{c}{Debiased} & \multicolumn{1}{c}{Original} & \multicolumn{1}{c}{Debiased} & \multicolumn{1}{c}{Original} & \multicolumn{1}{c}{Debiased} \\
     \midrule
     \underline{Accountant} & 64.77 (0.004) &63.27 (0.070) &63.45 (0.004) & \textbf{63.60} (0.068) & 1.32 (0.003) & \textbf{0.67} (0.005)\\
     Architect & 55.91 (0.016) & \textbf{58.40} (0.103) & 57.79 (0.012) & \textbf{64.94} (0.087) & 1.89 (0.009)& 6.55 (0.037) \\
     Attorney & 87.21 (0.001) & 86.37 (0.030) & 84.82 (0.004) & 84.39 (0.039)& 2.39 (0.003) & \textbf{1.99} (0.011)\\
     \underline{Chiropractor}$^{*}$ & 28.19 (0.004) & \textbf{29.07} (0.061)& 20.00 (0.004) & \textbf{25.85} (0.074) & 8.19 (0.004) & \textbf{3.38} (0.022) \\
     \underline{Comedian}$^*$ & 78.30 (0.006) & \textbf{80.11} (0.018)&62.31 (0.003) & \textbf{70.62} (0.047) &15.99 (0.004) & \textbf{9.49} (0.034)\\
     \underline{Composer} & 89.46 (0.002) & 87.21 (0.066) & 84.31 (0.003) & \textbf{84.47} (0.069) & 5.15 (0.001) & \textbf{2.74} (0.011)\\
     Dentist & 89.91 (0.001) & 89.63 (0.028) & 92.21 (0.001) & 92.01 (0.024) & 2.30 (0.002) & 2.38 (0.004) \\
     \underline{Dietitian}$^*$ & 56.25 (0.014) & \textbf{62.11} (0.043) & 76.20 (0.006) & \textbf{78.74} (0.064) & 19.95 (0.011) & \textbf{16.63} (0.022)\\
     Dj & 75.53 (0.011) & 69.67 (0.130) & 46.25 (0.014) & 43.50 (0.144) &29.28 (0.012)  & \textbf{26.17} (0.031)\\
     Filmmaker & 80.95 (0.004) & 76.46 (0.059) & 76.30 (0.006) & 75.68 (0.056) & 4.64 (0.004) & \textbf{1.12} (0.006)\\
     \underline{Interior Designer}& 44.23 (0.018) & \textbf{54.46} (0.086) & 68.67 (0.007) & 61.27 (0.076)  & 24.44 (0.012) & \textbf{6.81} (0.026) \\
     Journalist & 76.13 (0.003) & 72.45 (0.086) & 78.29 (0.004) & 73.92 (0.087) & 2.16 (0.003)& \textbf{1.46} (0.009) \\
     Model & 42.65 (0.002) & 42.60 (0.050) &81.35 (0.002)  & 76.34 (0.044) & 38.70 (0.002) & \textbf{33.74} (0.009) \\
     Nurse & 70.42 (0.004) & 67.67 (0.133) & 80.54 (0.005) & 75.53 (0.085) & 10.12 (0.006) & \textbf{7.85} (0.051)\\
     Painter & 77.47 (0.006) & 72.25 (0.120) & 79.39 (0.001) & 73.02 (0.136) & 1.92 (0.005) & \textbf{1.56} (0.008)\\
     \underline{Paralegal}$^*$ & 6.45 (0.000) &\textbf{18.39} (0.052)  & 35.29 (0.021) & \textbf{43.38} (0.087) & 28.84 (0.021) & \textbf{24.99} (0.040)\\
     Pastor & 62.09 (0.004) & 47.53 (0.078) & 55.78 (0.010) & 51.16 (0.079) & 6.31 (0.013) & \textbf{3.63} (0.005)\\
     Personal Trainer & 72.45 (0.009) & 68.06 (0.042) & 60.80 (0.012) & \textbf{66.05} (0.051) & 11.65 (0.011)& \textbf{3.44} (0.017)\\
     Photographer & 87.33 (0.001) & 84.78 (0.046) &84.25 (0.002) & 83.44 (0.054) & 3.08 (0.001) & \textbf{1.34} (0.009)\\
     Physician & 78.45 (0.008) & 74.57 (0.057) & 95.20 (0.001) &  94.79 (0.013)& 16.75 (0.007) & 20.22 (0.044)\\
     \underline{Poet}$^*$ & 74.92 (0.005) & \textbf{78.89} (0.029) & 76.44 (0.004) & \textbf{80.02} (0.029) & 1.52 (0.002) & \textbf{1.38} (0.014)\\
     Professor & 87.79 (0.003) & 86.93 (0.037) & 87.98 (0.002) & 86.84 (0.035) & 0.20 (0.001) & 0.45 (0.005)\\
     Psychologist & 62.30 (0.008) &  60.67 (0.049)& 65.32 (0.007) & 62.40 (0.054) & 3.02 (0.002)& \textbf{1.73} (0.006) \\
     \underline{Rapper} & 84.07 (0.011) & 77.80 (0.065) & 68.48 (0.031) & \textbf{74.78} (0.113) & 15.59 (0.023) & \textbf{5.48} (0.030)\\
     Software Engineer & 72.19 (0.014) & 66.49 (0.101) & 68.02 (0.022) & 66.50 (0.083) &4.17 (0.021) & \textbf{1.48} (0.015) \\
     \underline{Surgeon} & 62.36 (0.007) & 58.25 (0.093) & 36.12 (0.004) & \textbf{42.64} (0.094) & 26.24 (0.004) & \textbf{15.61} (0.005) \\
     Teacher & 45.39 (0.010) & 40.61 (0.102) & 58.98 (0.008) & 51.53 (0.104) & 13.59 (0.003)& \textbf{10.92} (0.004)\\
     \underline{Yoga Teacher}$^*$ & 43.36 (0.020) & \textbf{57.19} (0.053) & 67.59 (0.010) & \textbf{69.91} (0.051) & 24.23 (0.022) & \textbf{12.73} (0.016) \\
    \bottomrule
    \end{tabular}
    \end{adjustbox}

\end{table*}

\section{Prelim Experiment: Counterfactual Data Augmentation with Iterative Null Space Projection}

As an immediate possible next step, we experiment with the possibility of combining two existing methods to achieve a better satisfaction rate. We want to up weight protected group populations in the training data in the pre-processing step and use the augmented data as input to a debiasing layer \cite{zemel2013learning, woodworth2017learning}. In \Cref{male change}, \Cref{appendixC}, we evaluate the scale of changes with respect to the debiasing technique and profession groups. Across all the professions, CDA has the most large positive scale changes. We implement the counterfactual data augmentation technique on the input data of INLP (\Cref{cda-inlp-evaluation}). However, we do not see a huge improvement in combining the two existing debiasing techniques we evaluate in this work.

 \begin{table*}[ht]
     \caption{\textbf{Group-wise preliminary model performance}}
    \label{cda-inlp-evaluation}
    \centering
    \small
    \captionsetup{font=footnotesize}
    \begin{adjustbox}{width=\textwidth}
    \begin{tabular}{@{}lrrrrrr@{}}
    \toprule
     & \multicolumn{2}{c}{TPR Male} &   \multicolumn{2}{c}{TPR Female} & \multicolumn{2}{c}{GAP}\\
     \cmidrule(lr){2-3} \cmidrule(l){4-5} \cmidrule(l) {6-7}
    {Profession} & \multicolumn{1}{c}{Original} & \multicolumn{1}{c}{Debiased} & \multicolumn{1}{c}{Original} & \multicolumn{1}{c}{Debiased} & \multicolumn{1}{c}{Original} & \multicolumn{1}{c}{Debiased} \\
     \midrule
     \underline{Accountant}* & 64.77 (0.004) & \textbf{65.76} (0.033)&63.45 (0.004) & \textbf{65.72} (0.026) & 1.32 (0.003) & \textbf{0.85} (0.005)\\
     Architect & 55.91 (0.016) & 42.54 (0.113) & 57.79 (0.012) & 51.96 (0.084)& 1.89 (0.009)& 9.42 (0.045) \\
     Attorney & 87.21 (0.001) &  85.93 (0.039)& 84.82 (0.004) & 84.13 (0.039)& 2.39 (0.003) & \textbf{1.80} (0.003)\\
     \underline{Chiropractor} & 28.19 (0.004) & 26.34 (0.035) & 20.00 (0.004) & \textbf{23.49} (0.060) & 8.19 (0.004) & \textbf{2.86} (0.026) \\
     Comedian & 78.30 (0.006) & 72.55 (0.048) &62.31 (0.003) &61.38 (0.053) &15.99 (0.004) &\textbf{11.16} (0.018) \\
     Composer & 89.46 (0.002) & 82.59 (0.058) & 84.31 (0.003) & 80.43 (0.061) & 5.15 (0.001) & \textbf{2.17} (0.011)\\
     Dentist & 89.91 (0.001) & 87.91 (0.024) & 92.21 (0.001) & 90.28 (0.016) & 2.30 (0.002) & 2.36 (0.009) \\
     \underline{Dietitian}* & 56.25 (0.014) & \textbf{66.05} (0.011) & 76.20 (0.006) & \textbf{81.61} (0.024) & 19.95 (0.011) & \textbf{15.56} (0.014)\\
     \underline{Dj}* & 75.53 (0.011) & \textbf{77.16} (0.034) & 46.25 (0.014) & \textbf{48.50} (0.065) &29.28 (0.012)  & \textbf{28.66} (0.034)\\
     \underline{Filmmaker}* & 80.95 (0.004) & \textbf{83.37} (0.032) & 76.30 (0.006) & \textbf{81.30} (0.035) & 4.64 (0.004) & \textbf{2.07} (0.005)\\
     \underline{Interior Designer}& 44.23 (0.018) &\textbf{60.00} (0.139) & 68.67 (0.007) & 67.80 (0.134) & 24.44 (0.012) & \textbf{7.80} (0.029)\\
     Journalist & 76.13 (0.003) & 73.19 (0.037) & 78.29 (0.004) & 74.32 (0.042) & 2.16 (0.003)&\textbf{1.13} (0.008) \\
     \underline{Model} & 42.65 (0.002) & \textbf{43.13} (0.067) &81.35 (0.002)  & 77.82 (0.065) & 38.70 (0.002) & \textbf{34.69} (0.010)\\
     \underline{Nurse} & 70.42 (0.004) & \textbf{73.42} (0.028) & 80.54 (0.005) & 77.19 (0.034) & 10.12 (0.006) & \textbf{3.78} (0.008) \\
     Painter & 77.47 (0.006) & 76.73 (0.060) & 79.39 (0.001) & 78.99 (0.066) & 1.92 (0.005) & 2.26 (0.008)\\
     \underline{Paralegal}* & 6.45 (0.000) & \textbf{20.65} (0.024) & 35.29 (0.021) & \textbf{43.85} (0.085) & 28.84 (0.021) & \textbf{23.21} (0.063)\\
     Pastor & 62.09 (0.004) & 52.63 (0.099) & 55.78 (0.010) &51.56 (0.101) & 6.31 (0.013) & \textbf{2.69} (0.009)\\
     Personal Trainer & 72.45 (0.009) & 64.69 (0.033) & 60.80 (0.012) & 59.14 (0.040) & 11.65 (0.011)& \textbf{5.56} (0.019) \\
     Photographer & 87.33 (0.001) & 83.57 (0.065) &84.25 (0.002) & 80.55 (0.078) & 3.08 (0.001) & \textbf{3.02} (0.014)\\
     Physician & 78.45 (0.008) & \textbf{78.56} (0.045) & 95.20 (0.001) & \textbf{95.81} (0.007) & 16.75 (0.007) & 17.25 (0.038)  \\ 
     Poet & 74.92 (0.005) &  \textbf{75.69} (0.051)& 76.44 (0.004) & \textbf{77.96} (0.053) & 1.52 (0.002) & 2.27 (0.012)\\
     Professor & 87.79 (0.003) & 86.31 (0.030) & 87.98 (0.002) & 85.66 (0.038) & 0.20 (0.001) & 0.78 (0.009)\\
     Psychologist & 62.30 (0.008) &61.92 (0.060) & 65.32 (0.007) & 63.54 (0.067) & 3.02 (0.002)& \textbf{1.63} (0.008) \\
     \underline{Rapper} & 84.07 (0.011) & 79.27 (0.027) & 68.48 (0.031) & \textbf{82.61} (0.031) & 15.59 (0.023) & \textbf{3.75} (0.039)\\
     Software Engineer & 72.19 (0.014) & \textbf{80.18} (0.074) & 68.02 (0.022) & \textbf{75.94} (0.062) &4.17 (0.021) & 4.24 (0.015) \\
     \underline{Surgeon} & 62.36 (0.007) & 61.50 (0.056) & 36.12 (0.004) & \textbf{43.77} (0.044)& 26.24 (0.004) & \textbf{17.74} (0.015)\\
     Teacher & 45.39 (0.010) & 41.41 (0.060)& 58.98 (0.008) & 54.95 (0.060)& 13.59 (0.003)& \textbf{13.55} (0.008) \\
     \underline{Yoga Teacher} & 43.36 (0.020) & \textbf{56.25} (0.035) & 67.59 (0.010) & 66.38 (0.017) & 24.23 (0.022) & \textbf{10.13} (0.030) \\
    \bottomrule
    \end{tabular}
    \end{adjustbox}

\end{table*} 

\section{Evaluation Metrics}
\label{appendixB}

\textbf{TPR.} The True Positive Rate quantifies a system's proficiency in accurately identifying true positives within the overall population of the designated positive group(s)

\textbf{GAP.} GAP is closely related to the concept of fairness by Equality of Opportunities in \cite{hardt2016equality}, which introduces that if two individuals from a different group of the label are equally qualified for a positive outcome, they should have the same probability of being classified correctly.

Suppose we have one profession with a highly significant gap between males and females; by averaging group-wise TPR, the result would be affected by this specific significant gap while the result of TPR GAP with RMS will not by such case. 

\section{Group-wise Performance Changes}
\label{appendixC}

\begin{figure}[ht]
\centering
\subfloat{
    \begin{subfigure}{0.75\linewidth}
    % \centering
    \includegraphics[width=\linewidth]{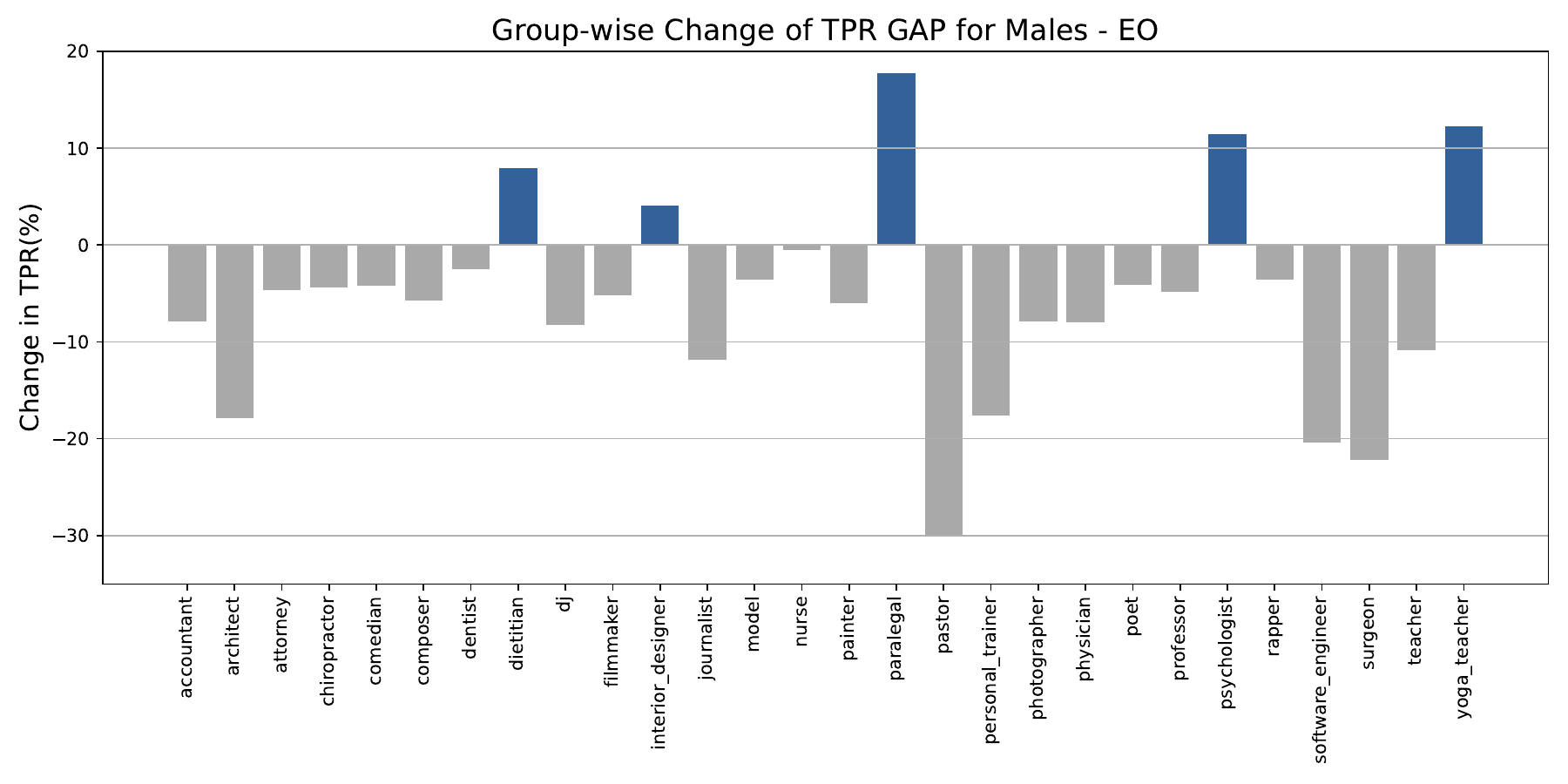}
    \end{subfigure} %
    }
    \\
\subfloat{
    \begin{subfigure}{0.75\linewidth}
    % \centering
    \includegraphics[width=\linewidth]{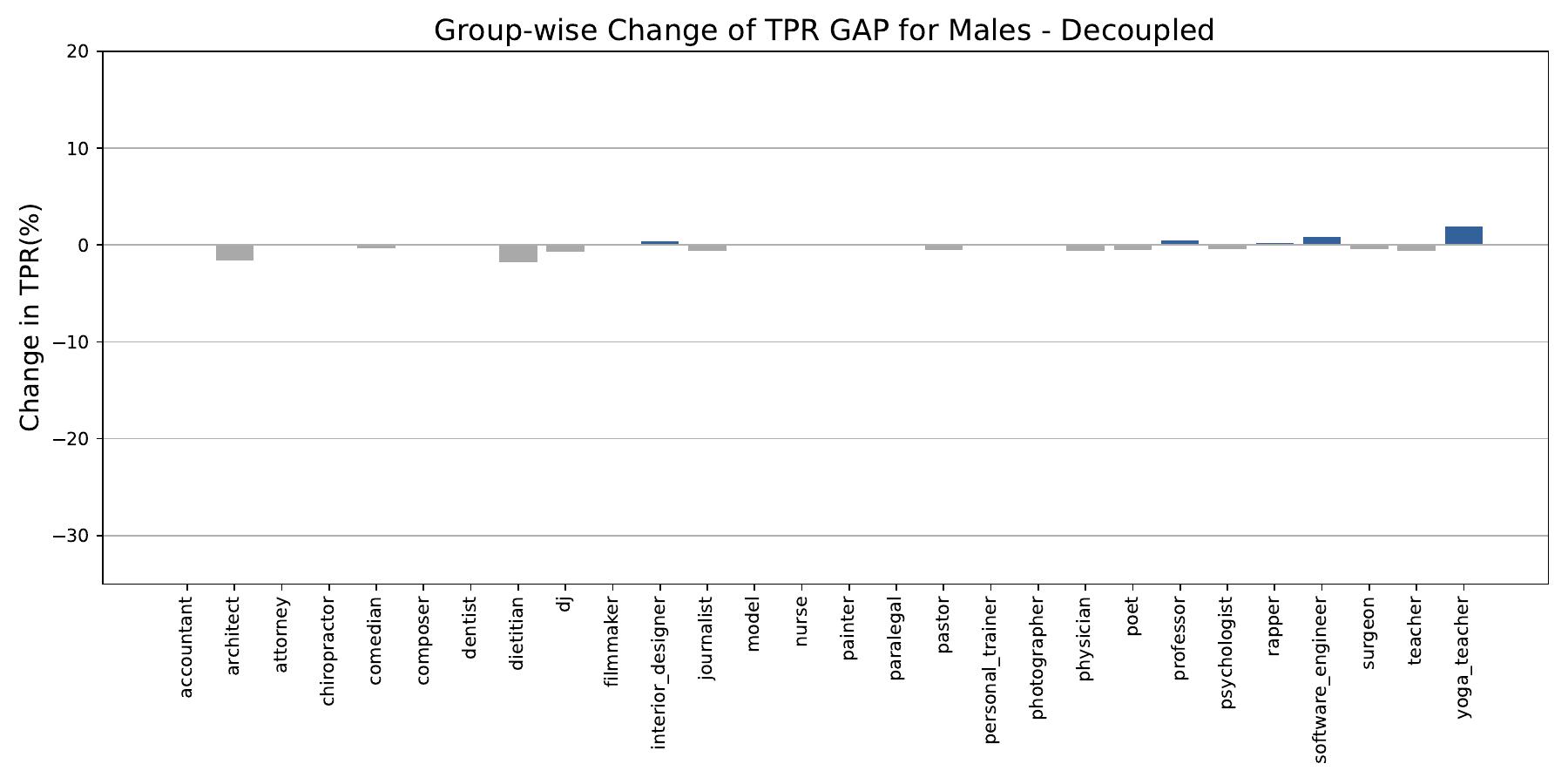}
    \end{subfigure}
}
\\
\subfloat{
    \begin{subfigure}{0.75\linewidth}
    % \centering
    \includegraphics[width=\linewidth]{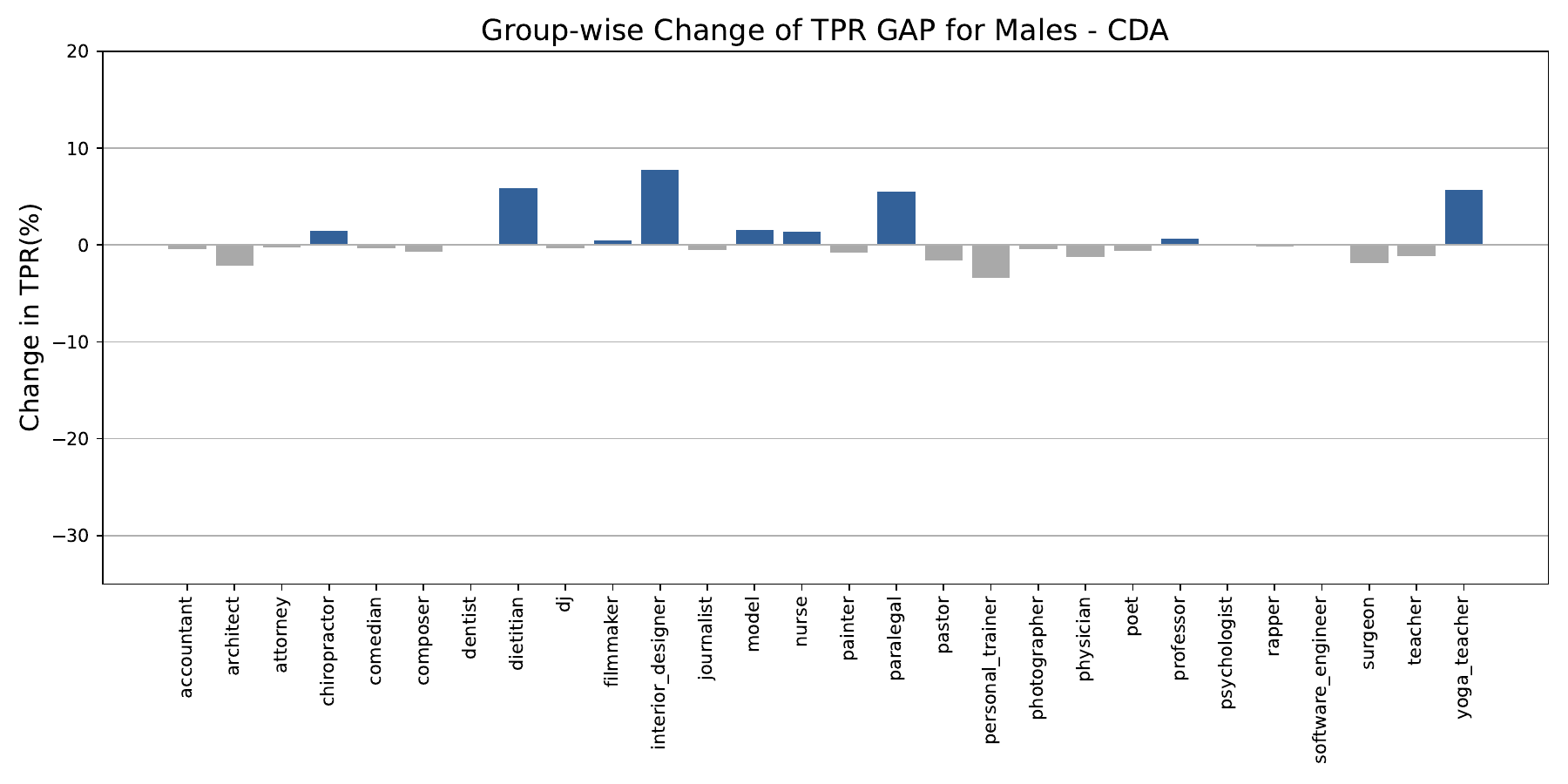}
    \end{subfigure} %
}
\\
\subfloat{
    \begin{subfigure}{0.75\linewidth}
    % \centering
    \includegraphics[width=\linewidth]{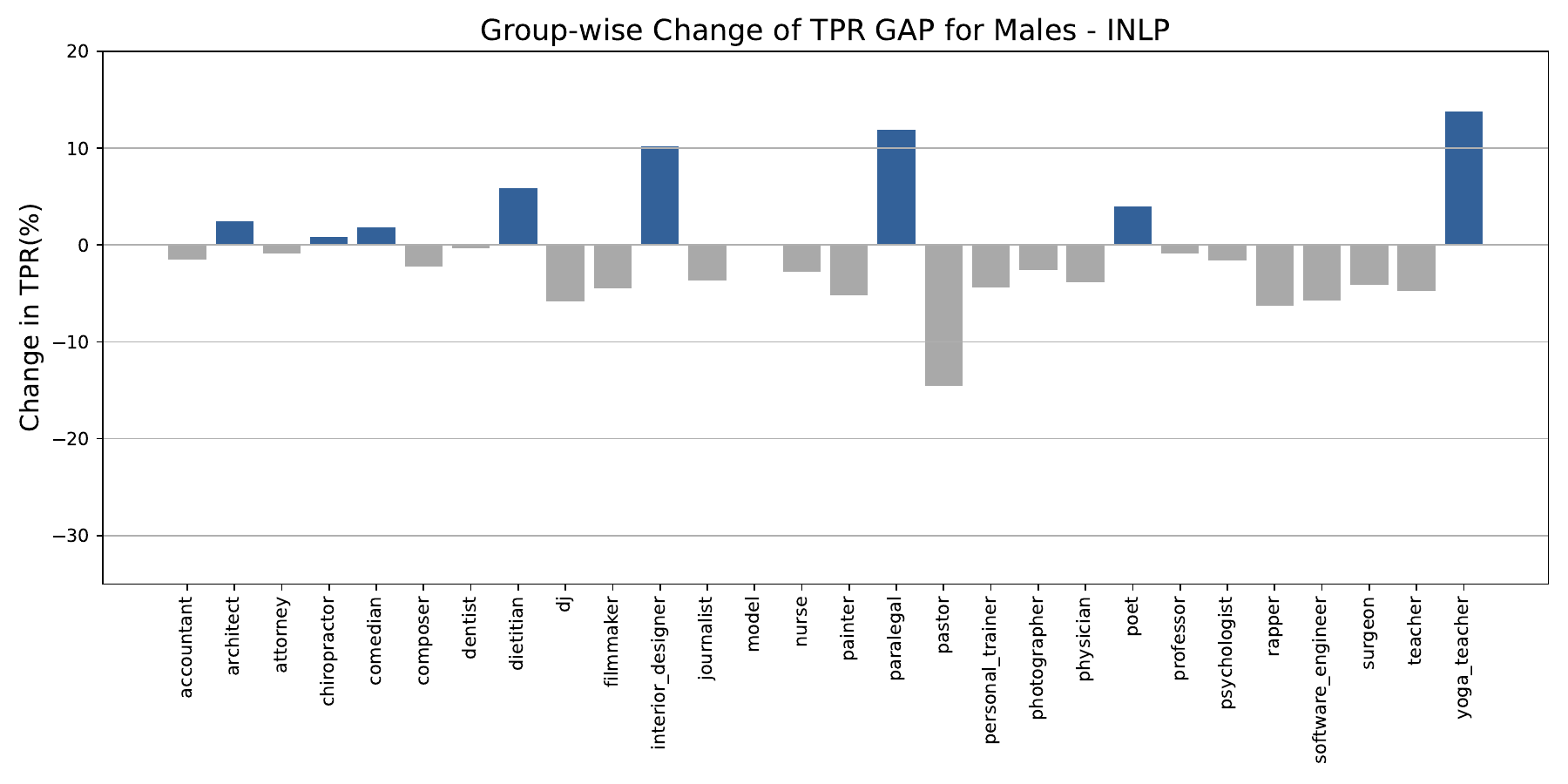}
    \end{subfigure} %
}
\caption{\textbf{Scale of changes in performance - TPR male}}
\label{male change}
\end{figure}

\begin{figure}[ht]
\centering
\subfloat{
    \begin{subfigure}{0.75\linewidth}
    % \centering
    \includegraphics[width=\linewidth]{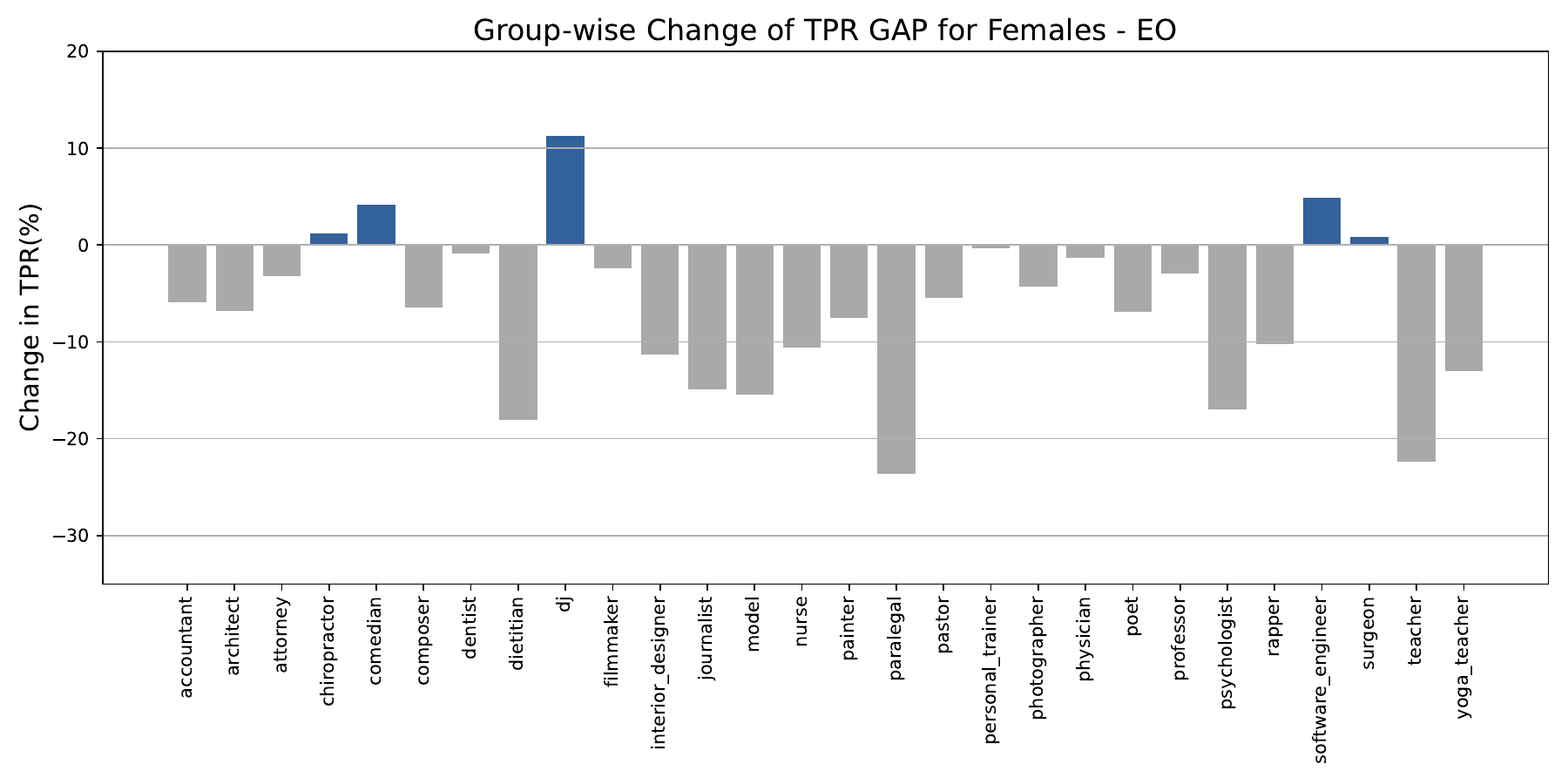}
    \end{subfigure} %
}
\\
\subfloat{
    \begin{subfigure}{0.75\linewidth}
    % \centering
    \includegraphics[width=\linewidth]{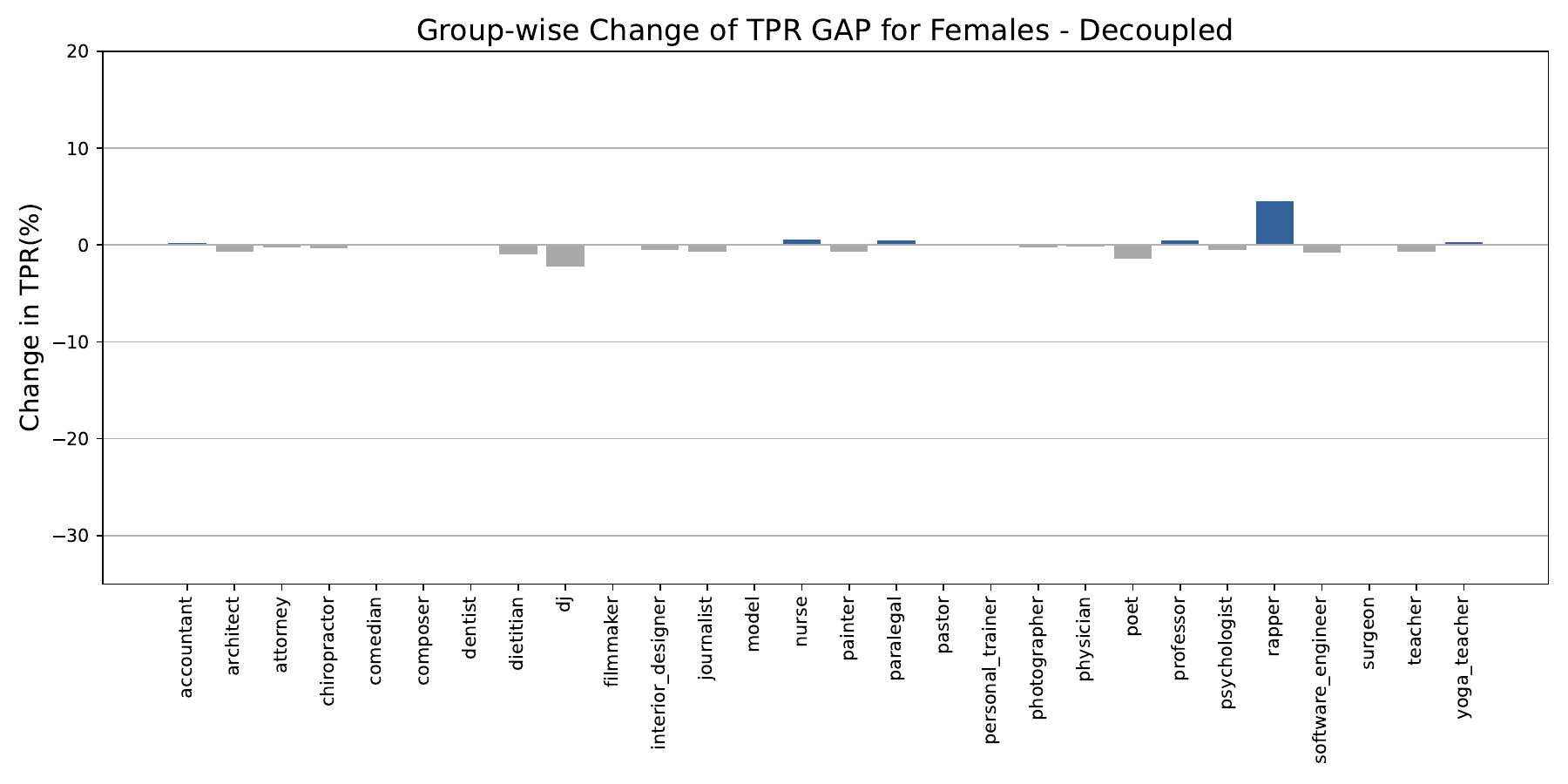}
    \end{subfigure}
}
\\
\subfloat{
\begin{subfigure}{0.75\linewidth}
    % \centering
    \includegraphics[width=\linewidth]{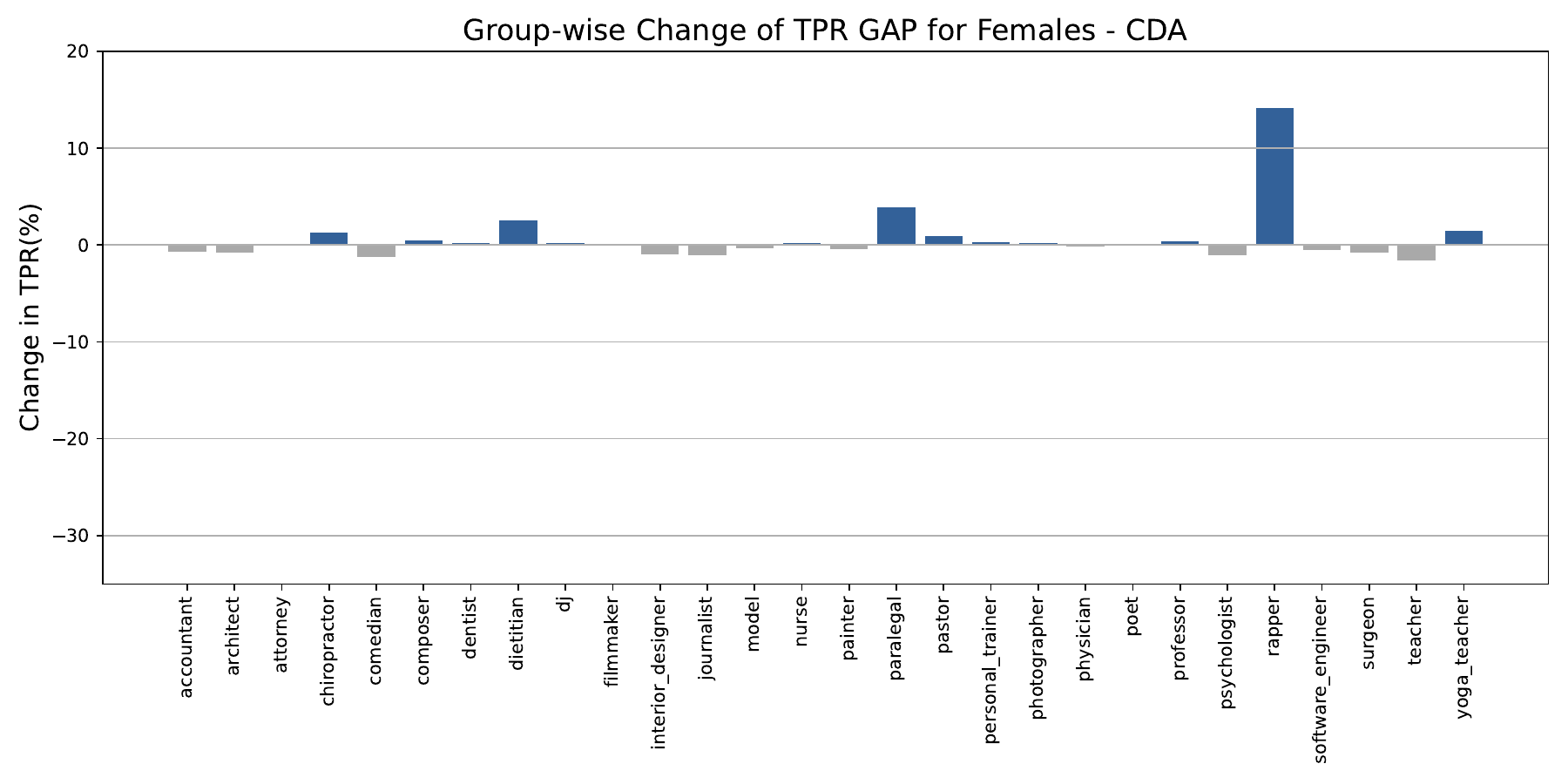}
\end{subfigure} %
}
\\
\subfloat{
\begin{subfigure}{0.75\linewidth}
    % \centering
    \includegraphics[width=\linewidth]{GWTPRF_INLP.pdf}
\end{subfigure} %
}
\caption{\textbf{Scale of changes in performance - TPR female}}
\label{female change}
\end{figure}

\begin{figure}[ht]
\centering
\subfloat{
    \begin{subfigure}{0.75\linewidth}
    % \centering
    \includegraphics[width=\linewidth]{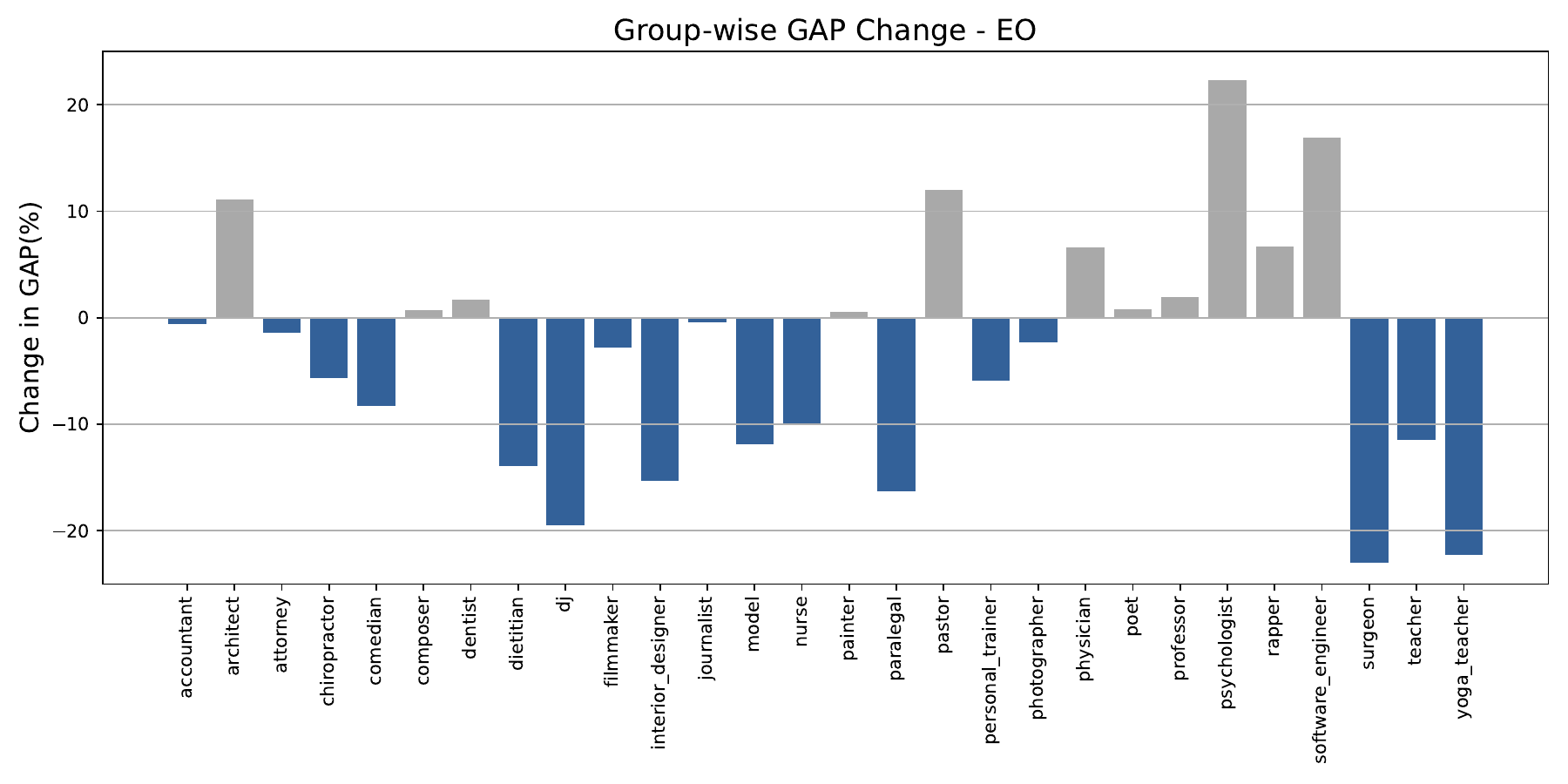}
    \end{subfigure} %
}
\\
\subfloat{
    \begin{subfigure}{0.75\linewidth}
    % \centering
    \includegraphics[width=\linewidth]{GWGAP_D.pdf}
    \end{subfigure}
}
\\
\subfloat{
\begin{subfigure}{0.75\linewidth}
    % \centering
    \includegraphics[width=\linewidth]{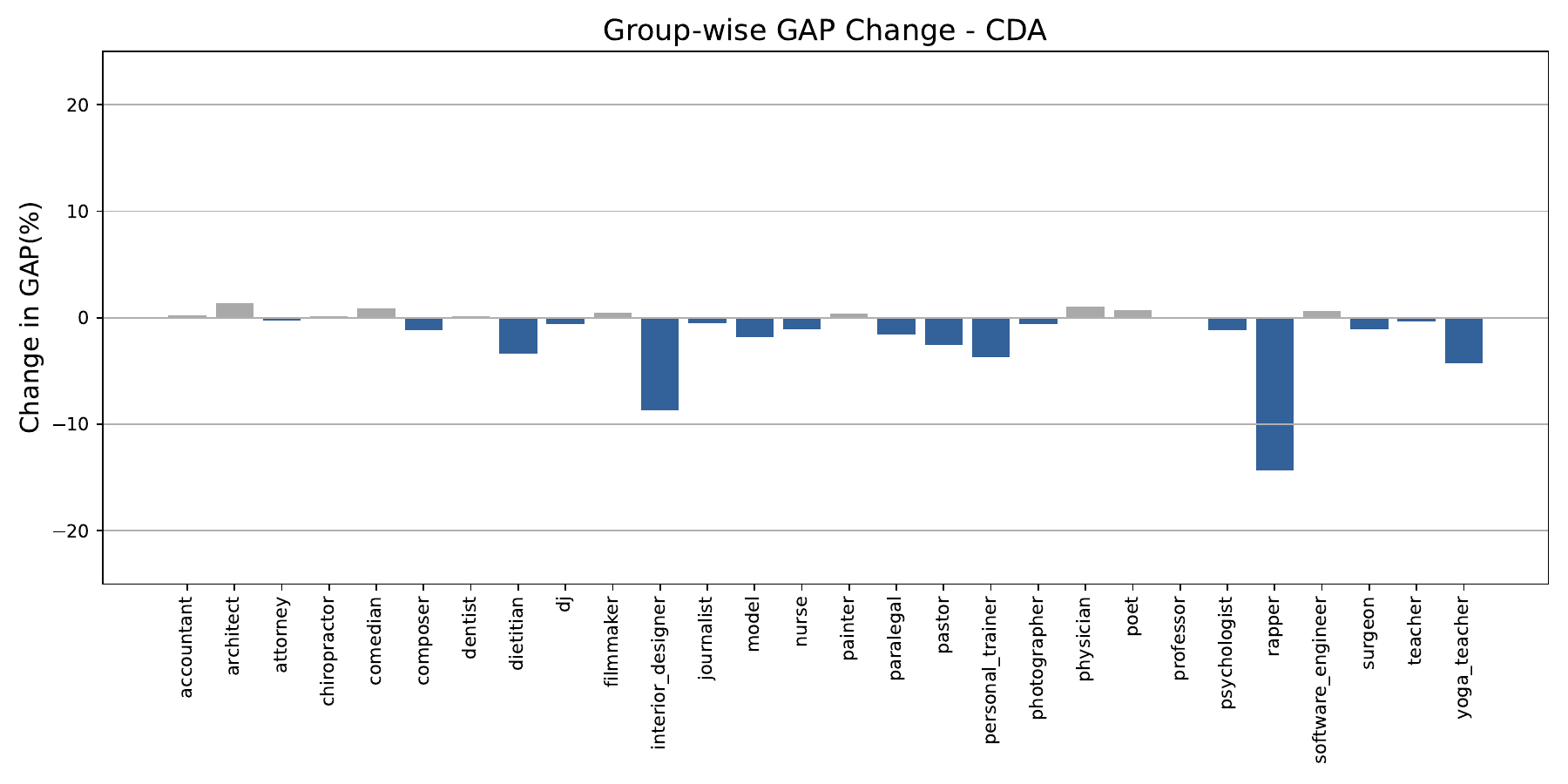}
\end{subfigure} %
}
\\
\subfloat{
\begin{subfigure}{0.75 \linewidth}
    \includegraphics[width=\linewidth]{{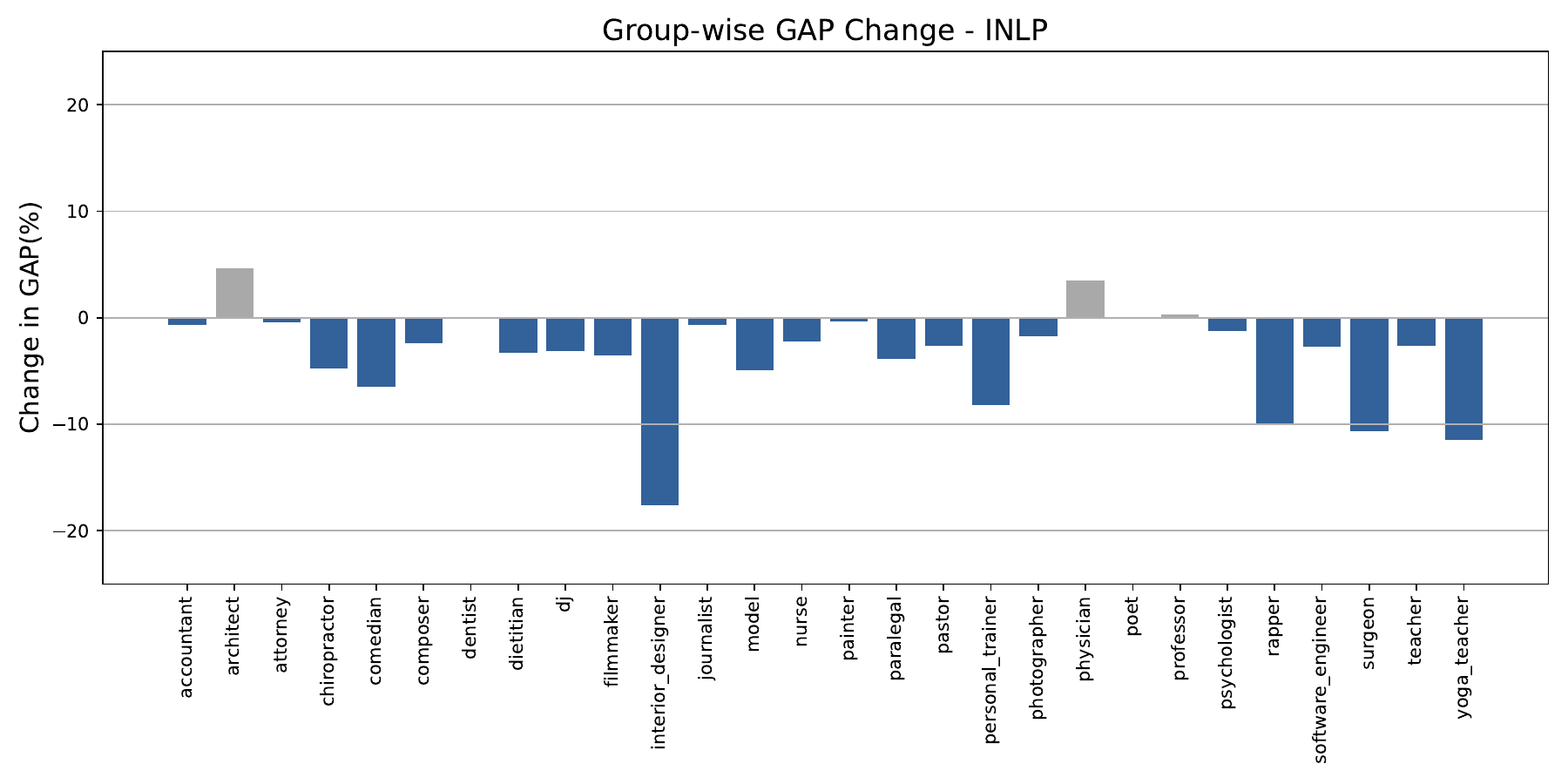}}
\end{subfigure}
}

\caption{\textbf{Scale of changes in performance - GAP}}
\label{gap change}
\end{figure}

We evaluate the changes in True Positive Rate (TPR) and the changes in \textsc{GAP} across all professions for males and females on the debiasing techniques (\Cref{male change}, \Cref{female change}, \Cref{gap change}). Through the figures, we calculate the change in model performance after the debiasing and before. The blue bars denote the directions in our interest, where we want an increase in the measure of TPR and a decrease in the measure of \textsc{GAP}.

\begin{figure}[ht]
\centering
\includegraphics[width=.9\linewidth]{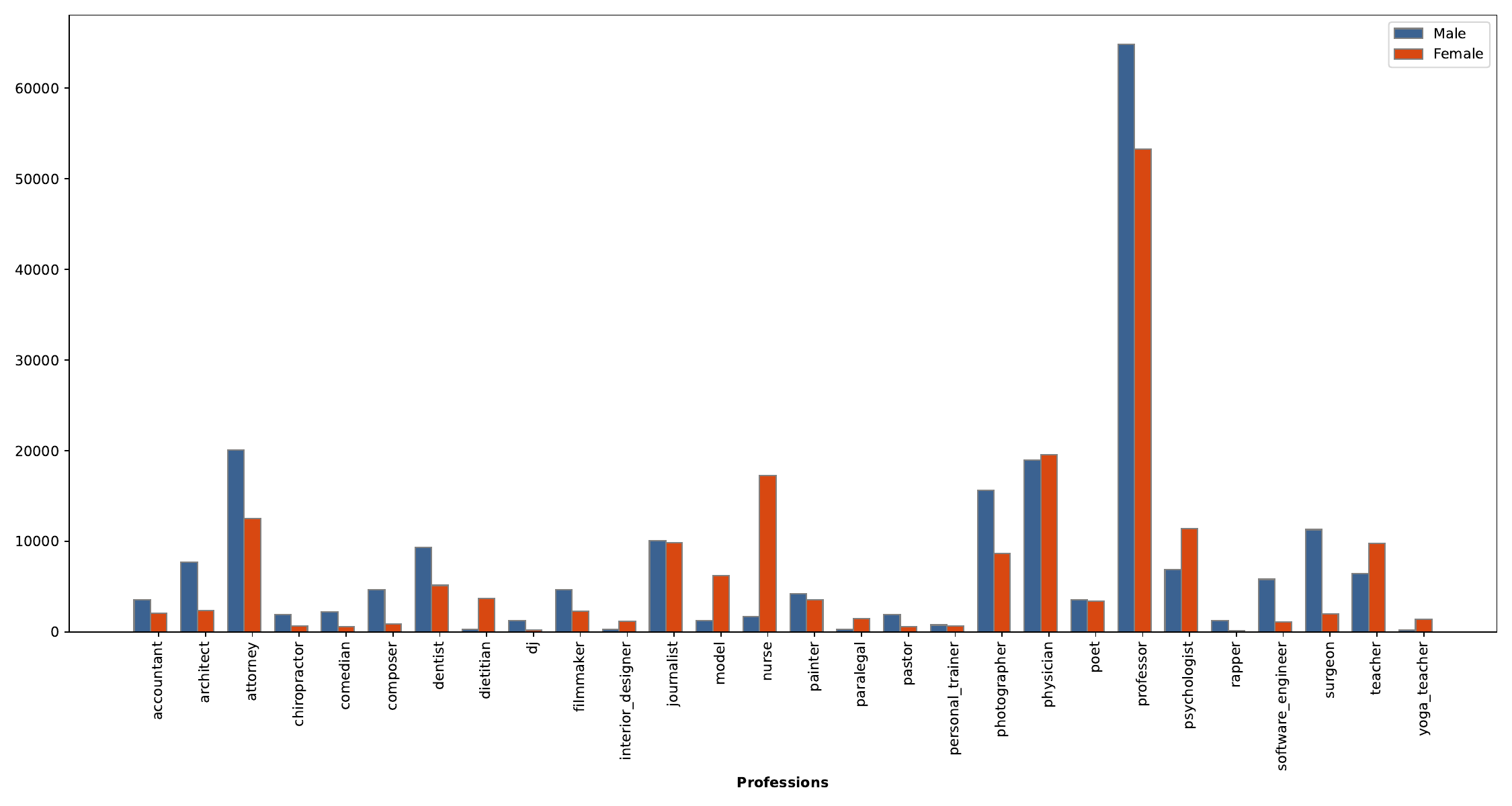}
\caption{\textbf{Bias in Bios group-wise study population}}
\label{study population}
\end{figure}

\section{Fairness with Awareness and Fairness with Unawareness}
\label{appendixD}
Debiasing approaches can be categorized as fairness with awareness and fairness with unawareness. Fairness with awareness refers to the approach where sensitive attributes, such as gender, race, or age, are explicitly considered in the model's debiasing process to ensure fair outcomes \cite{bolukbasi2016man, zhou-etal-2022-distantly, feng2021empowering}. This approach allows for targeted interventions to ensure that different demographic groups are treated equitably and has been used successfully in the fair classification literature \cite{hardt2016equality}. Fairness with unawareness refers to the approach where sensitive attributes are \textit{not} explicitly considered by the model \cite{guo-etal-2022-auto, elazar-goldberg-2018-adversarial, berg-etal-2022-prompt}. Instead, the model is designed to ensure fair outcomes without direct knowledge of the protected attributes. This approach helps maintain privacy and limits the potential for misuse and has been the main focus of prior work on debiasing language models. Still, it may not be as effective in addressing biases rooted in complex interactions between features or when there is a strong correlation between sensitive attributes and other input features. An example of fairness with unawareness technique is adversarial debiasing \cite{elazar-goldberg-2018-adversarial, berg-etal-2022-prompt}, which involves training a model to generate unbiased outputs while an adversary attempts to predict sensitive attributes from those outputs.

\section{Stopwords}
We remove certain commonly seen words from the biography.We borrowed the list of stopwords from \cite{ravfogel-etal-2020-null}:``i", ``me",``my", ``myself", ``we", ``our", ``ours", ``ourselves", ``you", ``your", ``yours", ``yourself", ``yourselves", ``he", ``him", ``his", ``himself", ``she", ``her", ``hers", ``herself", ``it", ``its", ``itself", ``they", ``them", ``their", ``theirs", ``themselves", ``what", ``which", ``who", ``whom", ``this", ``that", ``these", ``those", ``am", ``is", ``are", ``was", ``were", ``be", ``been", ``being", ``have", ``has", ``had", ``having", ``do", ``does", ``did", ``doing", ``a", ``an", ``the", ``and", ``but", ``if", ``or", ``because", ``as", ``until", ``while", ``of", ``at", ``by", ``for", ``with", ``about", ``against", ``between", ``into", ``through", ``during", ``before", ``after", ``above", ``below", ``to", ``from", ``up", ``down", ``in", ``out", ``on", ``off", ``over", ``under", ``again", ``further", ``then", ``once", ``here", ``there", ``when", ``where", ``why", ``how", ``all", ``any", ``both", ``each", ``few", ``more", ``most", ``other", ``some", ``such", ``no", ``nor", ``not", ``only", ``own", ``same", ``so", ``than", ``too", ``very", ``s", ``t", ``can", ``will", ``just", ``don", ``should", ``now"

\end{document}